\documentclass[letterpaper]{article}
\usepackage{uai2018}
\usepackage[margin=1in]{geometry}

\usepackage{times}

\usepackage{booktabs}
\usepackage{natbib}
\usepackage{amssymb}
\usepackage{amsmath}
\usepackage{graphicx}
\usepackage{subcaption}
\usepackage{xcolor} 
\usepackage{multicol}
\usepackage{multirow}
\usepackage{physics}
\usepackage{amsthm}
\usepackage{thmtools}

\DeclareMathOperator*{\KL}{\mathbb{KL}}

\definecolor{darkblue}{HTML}{0000AA}
\usepackage[colorlinks=true, linkcolor=darkblue, urlcolor=darkblue, citecolor=darkblue]{hyperref}

\newtheorem{proposition}{Proposition}

\declaretheoremstyle[%
 headfont=\normalfont\itshape,%
 qed=\qedsymbol%
]{mystyle}
\declaretheorem[name={Proof sketch},style=mystyle,unnumbered]{proofsketch}

\newcommand{\fw}{0.23\linewidth}

\title{Block Neural Autoregressive Flow}

\author{ {\bf Nicola De Cao} \\
University of Edinburgh\\
University of Amsterdam\\
{\tt nicola.decao@uva.nl}\\
\And
{\bf Wilker Aziz} \\
University of Amsterdam\\
{\tt w.aziz@uva.nl} \\
\And
{\bf Ivan Titov} \\
University of Edinburgh\\
University of Amsterdam\\
{\tt ititov@inf.ed.ac.uk} \\
}

\begin{document}

\maketitle

\begin{abstract}
Normalising flows (NFs) map two density functions via a differentiable bijection whose Jacobian determinant can be computed efficiently. Recently, as an alternative to hand-crafted bijections, \citet{huang2018neural} proposed neural autoregressive flow (NAF) which is a universal approximator for density functions. Their flow is a neural network (NN) whose parameters are predicted by another NN. The latter grows quadratically with the size of the former and thus an efficient technique for parametrization is needed. We propose \emph{block neural autoregressive flow} (B-NAF), a much more compact universal approximator of density functions, where we model a bijection directly using a single feed-forward network. Invertibility is ensured by carefully designing each affine transformation with block matrices that make the flow autoregressive and (strictly) monotone. We compare B-NAF to NAF and other established flows on density estimation and approximate inference for latent variable models. Our proposed flow is competitive across datasets while using orders of magnitude fewer parameters.
\end{abstract}

\section{INTRODUCTION}
Normalizing flows (NFs) map two probability density functions via an invertible transformation with tractable Jacobian \citep{tabak2010density}. They have been employed in contexts where we need to model a complex density while maintaining efficient sampling and/or density assessments. In density estimation, for example, NFs are used to map observations from a complex (and unknown) data distribution to samples from a simple base distribution \citep{rippel2013high}. In variational inference, NFs map from a simple fixed random source (e.g. a standard Gaussian) to a complex posterior approximation while allowing for reparametrized gradient estimates and density assessments \citep{rezende2015variational}.

Much of the research in NFs focuses on designing expressive transformations while satisfying practical constraints. In particular, autoregressive flows (AFs) decompose a joint distribution over $y \in \mathbb R^d$ into a product of $d$ univariate conditionals. A transformation $y=f(x)$, that realizes such a decomposition, has a lower triangular Jacobian with the determinant (necessary for application of the change of variables theorem for densities) computable in $O(d)$-time. \citet{kingma2016improved} proposed \emph{inverse autoregressive flows} (IAFs), an AF based on transforming each conditional by a composition of a finite number of trivially invertible affine transformations.

Recently, \citet{huang2018neural} introduced \emph{neural autoregressive flows} (NAFs). They replace IAF's transformation by a learned bijection realized as a strictly monotonic neural network. Notably, they prove that their method is a universal approximator of real and continuous distributions. Though, whereas parametrizing affine transformations in an IAF requires predicting $d$ pairs of scalars per step of the flow, parametrizing a NAF requires predicting all the parameters of a feed-forward \emph{transformer} network. The \emph{conditioner} network which parametrizes the transformer grows quadratically with the width of the transformer network, thus efficient parametrization techniques are necessary. A NAF is an instance of a hyper-network \citep{ha2016hypernetworks}. 

\paragraph{Contribution} 
We propose \emph{block neural autoregressive flows} (B-NAFs)
\footnote{\url{https://github.com/nicola-decao/BNAF}},
which are AFs based on a novel transformer network which transforms conditionals directly, i.e. without the need for a conditioner network. To do that we exploit the fact that invertibility only requires $\pdv*{y_i}{x_i} > 0$, and therefore, careful design of a feed-forward network can ensure that the transformation is both autoregressive (with unconstrained manipulation of $x_{<i}$) and strictly monotone (with positive $\pdv*{y_i}{x_i}$). We do so by organizing the weight matrices of dense layers in block matrices that independently transform subsets of the variables and constrain these blocks to guarantee that $\pdv*{y_i}{x_{j}} = 0$ for $j>i$ and that $\pdv*{y_i}{x_i} > 0$. Our B-NAFs are much more compact than NAFs while remaining universal approximators of density functions. We evaluate them both on density estimation and variational inference showing performance on par with state-of-the-art NFs, including NAFs, IAFs, and Sylvester flows \citep{berg2018sylvester}, while using orders of magnitude fewer parameters. 

\section{BACKGROUND}
In this section, we provide an introduction to normalizing flows and their applications (\S~\ref{sec:nf}). 
Then, we motivate autoregressive flows in \S~\ref{sec:auto} and present the necessary background for our contributions (\S~\ref{sec:naf}).

\subsection{NORMALIZING FLOW} \label{sec:nf}
A (finite) normalizing flow is a bijective function $f: \mathcal{X} \rightarrow \mathcal{Y}$ between two continuous random variables $X \in\mathcal{X}\subseteq \mathbb R^d $ and $Y \in \mathcal{Y} \subseteq \mathbb R^d$ \citep{tabak2010density}. The change of variables theorem expresses a relation between the probability density functions $p_Y(y)$ and $p_X(x)$:
\begin{equation} \label{eq:flow}
 p_Y(y) = p_X(x) \abs{ \det \mathbf{J}_{f(x)} }^{-1} \;,
\end{equation}
where $y = f(x)$, and $\abs{ \det \mathbf{J}_{f(x)} }$ is the absolute value of the determinant of the Jacobian of $f$ evaluated at $x$. The Jacobian matrix is defined as $\left(\mathbf{J}_{f(x)}\right)_{ij} = \pdv*{f(x)_i}{x_j}$. The determinant quantifies how $f$ locally expands or contracts regions of $\mathcal{X}$. Note that a composition of invertible functions remain invertible, thus a composition of NFs is itself a normalizing flow. 

\paragraph{Density estimation} 
In parametric density estimation, NFs map draws from complex distributions to draws from simple ones allowing assessments of a complex likelihood. Effectively, observations $x \sim p_{\text{data}}$ are modeled as draws from an NF $p_{X|\theta}$ whose parameters $\theta$ are chosen to minimize the Kullback-Leibler divergence $\KL(p_{\text{data}} \| p_{X|\theta})$ of the model $p_{X|\theta}$ from $p_{\text{data}}$: 
\begin{equation}\label{eq:NF-DE}
 \mathbb H(p_{\text{data}}) - \mathbb E_{\;p_{\text{data}}}\left[ \log p_Y(f(x)) \abs{\det \mathbf{J}_{f_\theta(x)}} \right] \; ,
\end{equation}
where $\mathbb H$ indicates the entropy. Minimizing such KL is equivalent to maximizing the log-likelihood of observations (see Appendix~\ref{app:kl_desity_estimation} for details of such derivation).

\paragraph{Variational inference} 
In variational inference for deep generative models, NFs map draws from a simple density $q_X$, e.g. $\mathcal N(0, I)$, to draws from a complex (multimodal) density $q_{Y|\theta}(y)$. The parameters $\theta$ of the flow are estimated to minimize the KL divergence $\KL(q_{Y|\theta} \| p_Y)$ of the true posterior $p_Y$ from $q_{Y|\theta}$, 
\begin{equation}\label{eq:NF-VI}
 \mathbb E_{\;q_X(x)} \left[\log \frac{q_X(x)\abs{\det \mathbf J_{f_\theta(x)}}^{-1}}{p_Y(f_\theta(x))}\right] ~,
\end{equation}
and note that this enables backpropagation through Monte Carlo (MC) estimates of the KL.

\paragraph{Tractability of NFs}
When optimizing normalizing flows with stochastic gradient descent, we have to evaluate a base density and compute the gradient with respect to its inputs. This poses no challenge as we have the flexibility to choose a simple density (e.g. uniform or Gaussian). In addition, for every $x$ (i.e. an observation in parametric density estimation (Equation~\ref{eq:NF-DE}) or a base sample in variational inference (Equation~\ref{eq:NF-VI})), the term $| \det \mathbf{J}_{f_\theta(x)} |$ has to be evaluated and differentiated with respect to the parameters $\theta$. Note that $f_\theta(x)$ is required to be invertible, as expressive as possible, and ideally fast to compute. In general, it is non-trivial to construct invertible transformations with efficiently computable $| \det \mathbf{J}_{f_\theta(x)} |$. Computing the determinant of a generic Jacobian $\mathbf{J}_{f_\theta(x)} \in \mathbb{R}^{d \times d}$ runs in $\mathcal{O}(d^3)$-time. 
Our work and current research on NFs aim at constructing parametrized flows which meet efficiency requirements while maximizing the expressiveness of the densities they can represent.

\subsection{AUTOREGRESSIVE FLOWS} \label{sec:auto} 
We can construct $f(x)$ such that its Jacobian is lower triangular, and thus has determinant $\prod_{i=1}^d \pdv*{f(x)_i}{x_i}$, which is computed in time $\mathcal O(d)$. Flows based on autoregressive transformations meet precisely this requirement \citep{kingma2016improved, pmlr-v80-oliva18a, huang2018neural}. For a multivariate random variable $X = \langle X_1, \ldots, X_d \rangle$ with $d>1$, we can use the chain rule to express the joint probability of $x$ as product of $d$ univariate conditional densities:
\begin{equation} \label{eq:chain}
 p_{X}(x) = p_{X_1}(x_1)\prod_{i=2}^d p_{X_i | X_{<i}} (x_i|x_{<i}) \; .
\end{equation}
When we then apply a normalizing flow to each univariate density, we have an autoregressive flow. 
Specifically, we can use a set of functions $f^{(i)}$ that can be decomposed via \emph{conditioners} $c^{(i)}$, and invertible \emph{transformers} $t^{(i)}$:
\begin{equation} \label{eq:conditioner}
 y_i = f_\theta^{(i)}(x_{\leq i}) = t_\theta^{(i)}(x_i, c_\theta^{(i)}(x_{<i})) \;,
\end{equation}
where each transformer $t^{(i)}$ must be an invertible function with respect to $x_i$, and each $c^{(i)}$ is an unrestricted function. The resulting flow has a lower triangular Jacobian since each $y_i$ depends only on $x_{\leq i}$. The flow is invertible when the Jacobian is constructed to have non-zero diagonal entries.

\subsection{NEURAL AUTOREGRESSIVE FLOW} \label{sec:naf}
\begin{figure}[t]
 \centering
 \begin{subfigure}[b]{0.95\linewidth}
 \centering
 \includegraphics[width=\linewidth]{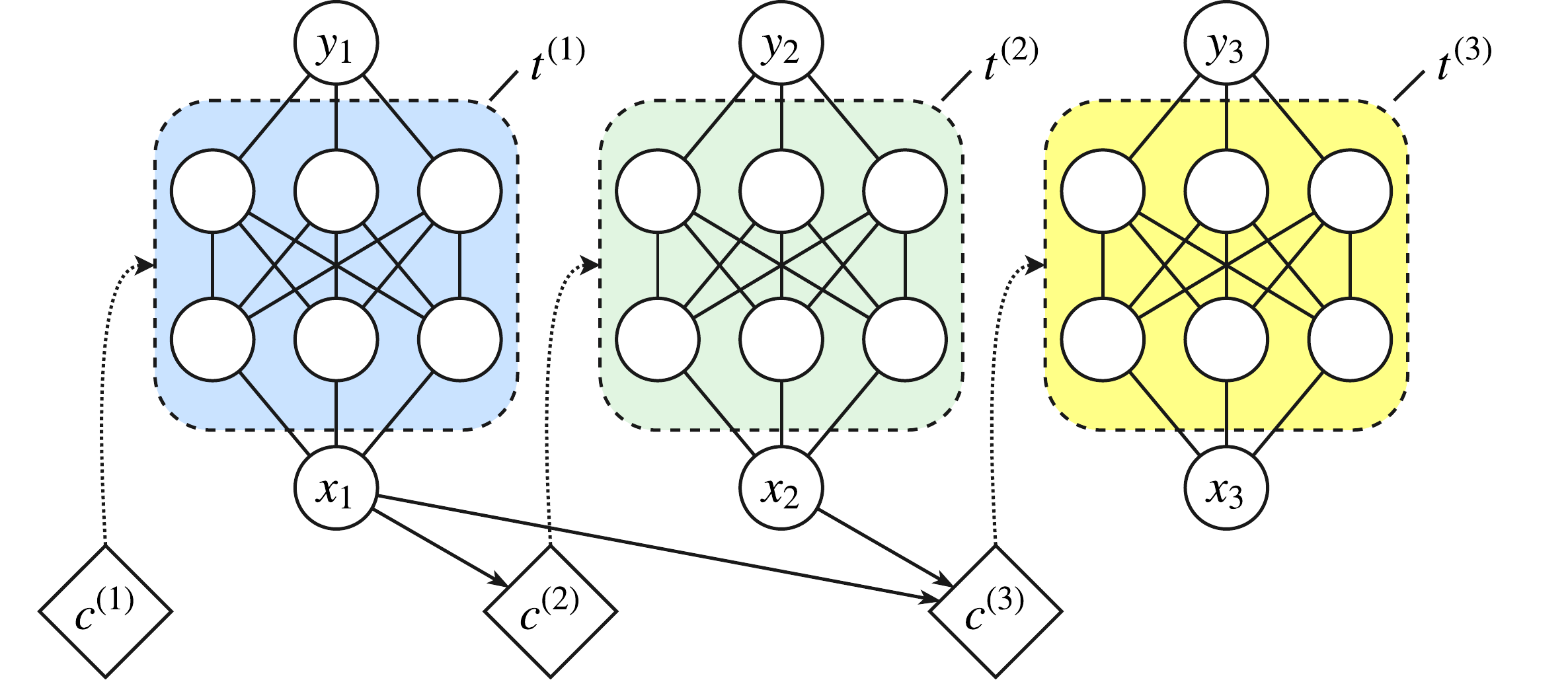}
 \caption{NAF: each $c^{(i)}$ is a neural network that predicts pseudo-parameters for $t^{(i)}$, which in turns processes $x_i$.}
 \label{fig:architectures_naf}
 \end{subfigure}%
 \vspace{1em}
 \begin{subfigure}[b]{\linewidth}
 \centering
 \includegraphics[width=0.8\linewidth]{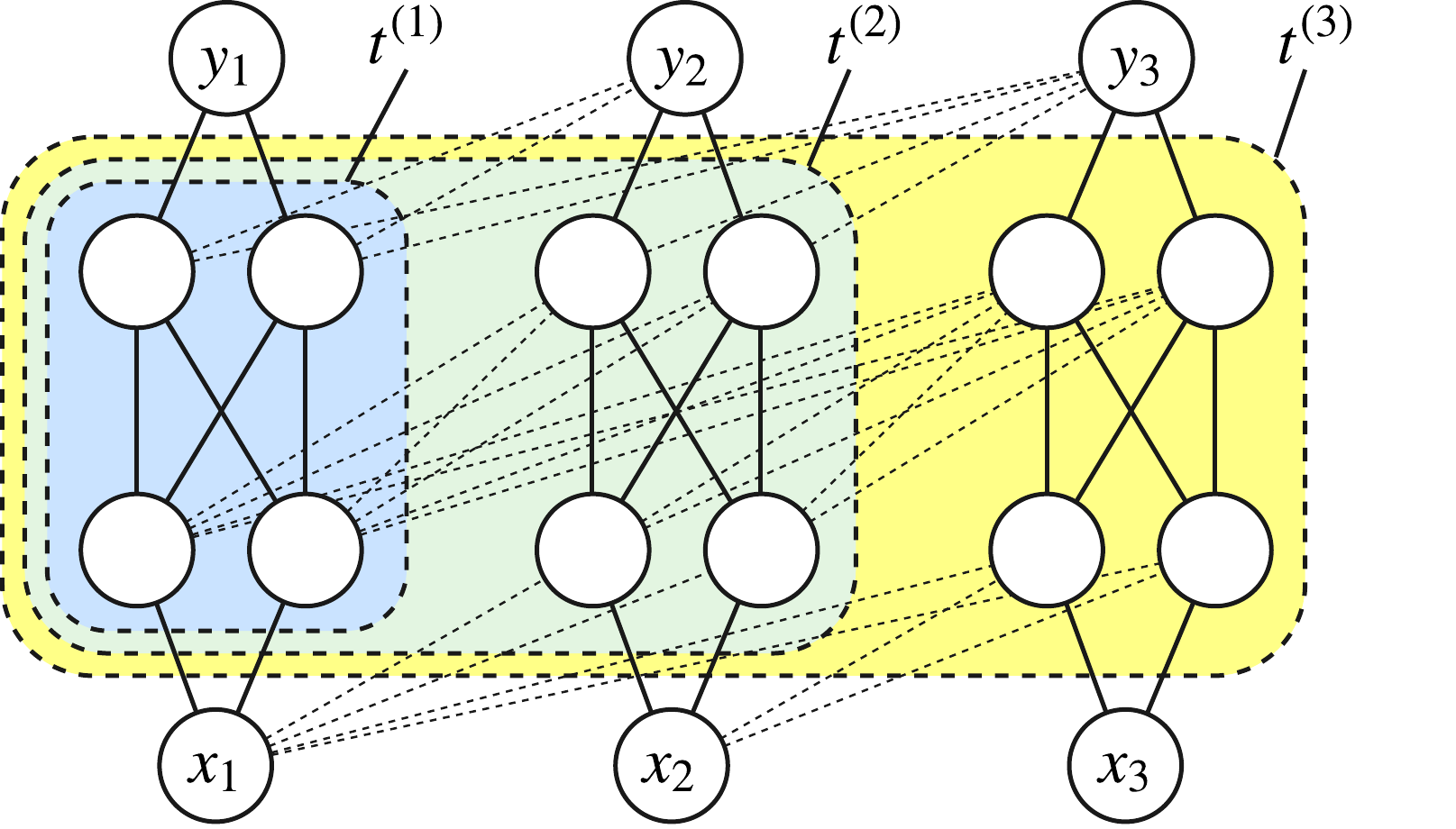}
 \caption{Our B-NAF: we do not use conditioner networks, instead we learn the flow network directly. Some weights are strictly positive (solid lines), others have no constraints (dashed lines).}
 \label{fig:architectures_bnaf}
 \end{subfigure}%
 \caption{Main differences between NAF \citep{huang2018neural} and our B-NAF. Both networks are autoregressive and invertible since $y_i$ is processed with a function $t^{(i)}$ which is monotonically increasing with respect to $x_i$ and there is an arbitrary dependence on $x_{<i}$.}
 \label{fig:architectures}
\end{figure}

The invertibility of the flow as a whole depends on each $t^{(i)}$ being an invertible function of $x_i$. 
For example, \citet{dinh2014nice} and \citet{kingma2016improved} model each $t^{(i)}$ as an affine transformation whose parameters are predicted by $c^{(i)}$. As argued by \citet{huang2018neural}, these transformations were constructed to be trivially invertible, but their simplicity leads to a cap on expressiveness of $f$, thus requiring complex conditioners and a composition of multiple flows. They propose instead to learn a complex bijection using a neural network monotonic in $x_i$ --- this only requires constraining $t^{(i)}$ to having non-negative weights and using strictly increasing activation functions. Figure~\ref{fig:architectures_naf} outlines a NAF. Each conditioner $c^{(i)}$ is an unrestricted function of $x_{<i}$. To parametrize a monotonically increasing transformer network $t^{(i)}$, the outputs of each conditioner $c^{(i)}$ are mapped to the positive real coordinate space by application of an appropriate activation (e.g. $\exp$). The result is a flexible transformation with lower triangular Jacobian whose diagonal elements are positive.\footnote{Note that the expressiveness of a NAF comes at the cost of analytic invertibility, i.e. though each $t^{(i)}$ is bijective, thus invertible in principle, inverting the network itself is non-trivial.}

For efficient computation of all pseudo-parameters, as \citet{huang2018neural} call the conditioners' outputs, they use a masked autoregressive network \citep{germain2015made}. The Jacobian of a NAF is computed using the chain rule on $f_\theta$ through all its hidden layers $\{h^{(\ell)}\}_{\ell=1}^l$:
\begin{equation} \label{eq:backprop}
 \mathbf{J}_{f_\theta(x)} = \left[ \grad_{h^{(l)}} y^{\color{white}(} \right] \left[ \grad_{h^{(l-1)}} h^{(l)} \right] \dots \left[ \grad_{x} h^{(1)} \right] \;.
\end{equation}
Since $f_\theta$ is autoregressive, $\mathbf{J}_{f_\theta(x)}$ is lower triangular and only the diagonal needs to be computed, i.e. $\pdv*{y_i}{x_i}$ for each $i$. Thus, this operation requires only computing the derivatives of each $t^{(i)}$, reducing the time complexity.

Because the universal approximation theorem for densities holds for NAFs \citep{huang2018neural}, increasing the expressiveness of a NAF is only a matter of employing larger transformer networks. However, the conditioner grows quadratically with the size of the transformer network and a combination of restricting the size of the transformer and a technique similar to \emph{conditional weight normalization} \citep{krueger2017bayesian} is necessary to reduce the number of parameters. In \S\ref{sec:bnaf} we propose to parametrize the transformer network directly, i.e. without a conditioner network, by exploiting the fact that the monotonicity constraint only requires $\pdv*{y_i}{x_i} > 0$, and therefore, careful design of a single feed-forward network can directly realize a transformation that is both autoregressive (with unconstrained manipulation of $x_{<i}$) and strictly monotone (with positive $\pdv*{y_i}{x_i}$). 

\section{BLOCK NEURAL AUTOREGRESSIVE FLOW} \label{sec:bnaf}
In the spirit of NAFs, we model each $f_\theta^{(i)}(x_{\le i})$
as a neural network with parameters $\theta$, but differently from NAFs, we do not predict $\theta$ using a conditioner network, and instead, we learn $\theta$ directly. In dense layers of $f^{(i)}_\theta$, we employ affine transformations with strictly positive weights to process $x_i$. This ensures strict monotonicity and thus invertibility of each $f_\theta^{(i)}$ with respect to $x_i$. However, we do not impose this constraint on affine transformations of $x_{<i}$. Additionally, we need to always use invertible activation functions to ensure that the whole network is bijective (e.g., $\tanh$ or LeakyReLU). Each $f_\theta^{(i)}$ is then a univariate flow implemented as an arbitrarily wide and deep neural network which can approximate any invertible transformation. Much like other AFs, we can efficiently compute all $f_\theta^{(i)}$ in parallel by employing a single masked autoregressive network \citep{germain2015made}. In the next section, we show how to construct each affine transformation using block matrices. From now on, we will refer to our novel family of flows as \emph{block neural autoregressive flows} (B-NAFs).

\subsection{AFFINE TRANSFORMATIONS WITH BLOCK MATRICES}
For each affine transformation of $x$, we parametrize the bias term freely and we construct the weight matrix $W \in \mathbb{R}^{a d \times b d}$ as a lower triangular \emph{block} matrix for some $a,b \geq 1$. We use $d \times (d + 1) / 2$ blocks $B_{ij} \in \mathbb{R}^{a \times b}$ for $i \in \{1,..,d\}$ and $1 \leq j \leq i$. We let $B_{ij}$ (with $i > j$) be freely parametrized and constrain diagonal blocks to be strictly positive applying an element-wise function $g: \mathbb{R} \rightarrow \mathbb{R}_{>0}$ to each of them. Thus:
\begin{equation} \label{eq:matrix}
 W=
 \begin{bmatrix}
 g(B_{11}) & 0 & \dots & 0 \\
 B_{21} & g(B_{22}) & \dots & 0 \\
 \vdots & \vdots & \ddots & \vdots \\
 B_{d1} & B_{d2} & \dots & g(B_{dd})
 \end{bmatrix} \;,
\end{equation}
where we chose $g(\cdot) = \exp(\cdot)$. Since the flow has to preserve the input dimensionality, the first and the last affine transformations in the network must have $b=1$ and $a=1$, respectively. Inside the network, the size of $a$ and $b$ can grow arbitrarily. 

The intuition behind the construction of $W$ is that every row of blocks $B_{i1},..,B_{ii}$ is a set of affine transformations (projections) that are processing $x_{\leq i}$. In particular, blocks in the upper triangular part of $W$ are set to zero to make the flow autoregressive. Since the blocks $B_{ii}$ are mapped to $\mathbb{R}_{>0}$ through $g$, each transformation in such set is strictly monotonic for $x_i$ and unconstrained on $x_{<i}$.

\paragraph{B-NAF with masked networks}
In practice, a more convenient parameterization of $W$ consists of using a full matrix $\hat{W} \in \mathbb{R}^{a d \times b d}$ which is then transformed applying two \emph{masking} operations. One mask $M_d \in \{0, 1\}^{a d \times b d}$ selects only elements in the diagonal blocks, and a second one $M_o$ selects only off-diagonal and lower diagonal blocks. Thus, for each layer $\ell$ we get
\begin{equation}
\label{eq:w}
 W^{(\ell)} = g\left( \hat{W}^{(\ell)} \right) \odot M^{(\ell)}_d + \hat{W}^{(\ell)} \odot M^{(\ell)}_o \;,
\end{equation}
where $\odot$ is the element-wise product. Figure~\ref{fig:architectures_bnaf} shows an outline of our block neural autoregressive flow.

Since each weight matrix $W^{(\ell)}$ has some strictly positive and some zero entries, we need to take care of a proper initialization which should take that into account. Indeed, weights are usually initialized to have a zero centred normally distributed output with variance dependent on the output dimensionality \citep{glorot2010understanding}. Instead of carefully designing a new initialization technique to take care of this, we choose to initialize all blocks with a simple distribution and to apply weight normalization \citep{salimans2016weight} to better cope the effect of such initialization. See Appendix~\ref{app:weight} for more details.

When constructing a stacked flow though a composition of $n$ B-NAF transformations, we add gated residual connections for improving stability such that the composition is $\hat f_n \circ \dots \circ \hat f_2 \circ \hat f_1$ where $\hat f_i (x) = \alpha f_i(x) + (1 - \alpha) x$ and $\alpha \in (0, 1)$ is a trainable scalar parameter.

\subsection{AUTOREGRESSIVENESS AND INVERTIBILITY}
In this section, we show that our flow $f_\theta: \mathbb{R}^d \rightarrow \mathbb{R}^d$ meets the following requirements: i) its Jacobian $\mathbf{J}_{f_\theta(x)}$ is lower triangular (needed for efficiency in computing its determinant), and ii) the diagonal entries of such Jacobian are positive (to ensure that $f_\theta$ is a bijection).

\begin{proposition} \label{pro:lower}
The final Jacobian $\mathbf{J}_{f_\theta(x)}$ of such transformation is lower triangular.
\end{proposition}
\begin{proofsketch}
When applying the chain rule (Equation~\ref{eq:backprop}), the Jacobian of each affine transformation is $W$ (Equation~(\ref{eq:w}), a lower triangular block matrix), whereas the Jacobian of each element-wise activation function is a diagonal matrix. A matrix multiplication between a lower triangular block matrix and a diagonal matrix yields a lower triangular block matrix, and a multiplication between two lower triangular block matrices results in a lower triangular block matrix. Therefore, after multiplying all matrices in the chain, the overall Jacobian is lower triangular.
\end{proofsketch}

\begin{proposition} \label{pro:positive}
When using strictly increasing activation functions (e.g., $\tanh$ or LeakyReLU), the diagonal entries of $\mathbf{J}_{f_\theta(x)}$ are strictly positive.
\end{proposition}
\begin{proofsketch}
When applying the chain rule (Equation~\ref{eq:backprop}), the Jacobian of each affine transformation has strictly positive values in its diagonal blocks where the Jacobian of each element-wise activation function is a diagonal matrix with strictly positive elements. When using matrix multiplication between two lower triangular block matrices (or one diagonal and one lower triangular block matrix) $C = A B$ the resulting blocks on the diagonal of $C$ are the result of a multiplication between only diagonal blocks of $A,B$. Indeed, such resulting blocks depend only on blocks of the same row and column partition. Using the notation of Equation~\ref{eq:matrix}, the resulting diagonal blocks of $C$ are $B^{(C)}_{ii} = g(B^{(A)}_{ii}) g(B^{(B)}_{ii})$. Therefore, they are always positive. Eventually, using the chain rule, the final Jacobian is a lower triangular matrix with strictly positive elements in its diagonal.
\end{proofsketch}

\subsection{EFFICIENT JACOBIAN COMPUTATION} \label{sec:bnaf_jacobian}
Proposition~\ref{pro:lower} is particularly useful for an efficient computation of $\det \mathbf{J}_{f_\theta(x)}$ since we only need the product of its diagonal elements $\pdv*{y_i}{x_i} $. Thus, we can avoid computing the other entries. Since the determinant is the result of a product of positive values, we also remove the absolute-value operation resulting in 
\begin{equation}\label{eq:backprop_log}
 \log |\det \mathbf{J}_{f_\theta(x)}| = \sum_{i=0}^d \log \left(\mathbf{J}_{f_\theta(x)} \right)_{ii} \;.
\end{equation}
Additionally, as per Proposition~\ref{pro:positive}, when using matrix multiplication, elements in the diagonal blocks (or entries) depend only on diagonal blocks of the same row and column partition. Since all diagonal blocks/entries are positive, we compute them directly in the log-domain to have more numerically stable operations:
\begin{equation}
 \begin{aligned}
 \log &\left(\mathbf{J}_{f_\theta(x)} \right)_{ii} = \log g(B^{(\ell)}_{ii})~\star \\
 &\log \mathbf{J}_{\sigma^{(\ell)}(h_\alpha^{(\ell-1)})}~\star \dots \star \log g(B_{ii}^{(1)}) \;,
 \end{aligned}
\end{equation}
where $\star$ denotes the log-matrix multiplication, $\sigma^{(\ell)}$ the strictly increasing non-linear activation function at layer $\ell$, and $\alpha$ indicates the set of indices corresponding to diagonal elements that depend on $x_i$. Notice that, since we chose $g(\cdot) = \exp(\cdot)$ we can remove all redundant operations $\log g (\cdot)$. The log-matrix multiplication $C = A \star B$ of two matrices $A \in \mathbb{R}^{m \times n} = \log \hat{A}$ and $B \in \mathbb{R}^{n \times p} = \log \hat{B}$ can be implemented with a stable \emph{log-sum-exp} operation since
\begin{equation}
 C_{ij} = \log \sum_{k=1}^n \exp \left( \hat{A}_{ik} + \hat{B}_{kj} \right) \;.
\end{equation}
A similar idea is also employed in NAF.

\subsection{UNIVERSAL DENSITY APPROXIMATOR}
In this section, we expose an intuitive proof sketch that our block neural autoregressive flow can approximate any real continuous probability density function (PDF).

Given a multivariate real and continuous random variable $X= \langle X_1, \ldots, X_d \rangle$, its joint distribution can be factorized into a set of univariate conditional distributions (as in Equation~\ref{eq:chain}), using an arbitrary ordering of the variables, and we can define a set of univariate conditional cumulative distribution functions (CDFs) $Y_i = F_{X_i|X_{<i}}(x_i|x_{<i}) = \mathbb P[X_i \leq x_i|X_{<i}=x_{<i}]$. According to \citet{hyvarinen1999nonlinear}, such decomposition exists and each individual $Y_i$ is independent as well as uniformly distributed in $[0, 1]$. Therefore, we can see $F_X$ as a particular \emph{normalizing flow} that maps $X \in \mathbb{R}^n$ to $Y \in [0, 1]^n$ where the distribution $p_Y$ is uniform in the hyper-cube $[0, 1]^n$. Note that $F_X$ is an autoregressive function and its Jacobian has a positive diagonal since $\partial y_i / \partial x_i = p_{X_i|X_{<i}}(x_i|x_{<i})$.

If each univariate flow $f_\theta^{(i)}$ (see Equation~\ref{eq:conditioner}) can approximate any invertible univariate conditional CDF, then $f_\theta$ can approximate any PDF \citep{huang2018neural}. Note that in general, a CDF $F_{X_i|X_{<i}}$ is non-decreasing, thus not necessary invertible \citep{park2018fundamentals}. Using B-NAF, each CDF is approximated with an arbitrarily large neural network and the output can be eventually mapped to $(0,1)$ with a sigmoidal function. Recalling that we only use positive weights for processing $x_i$, a neural network with non-negative weights is an universal approximator of monotonic functions \citep{daniels2010monotone}. We use \emph{strictly} positive weights to approximate a \emph{strictly} monotonic function for $x_i$ and we use arbitrary weights for $x_{<i}$ (as there is no monotonicity constraint for them). Therefore, B-NAF can approximate any invertible CDF, and thus its corresponding PDF.

\section{RELATED WORK}
Current research on NFs focuses on constructing expressive parametrized invertible trasformations with tractable Jacobians. \citet{rezende2015variational} were the first to suggest the use of parameterized flows in the context of variational inference proposing two parametric families: the planar and the radial flow. A drawback and bottleneck of such flows is that their power comes from stacking a large number of such transformations. More recently, \citet{berg2018sylvester} generalized the use of planar flows showing improvements without increasing the number of transformations, and instead, by making each transformation more expressive.

In the context of density estimation, \citet{germain2015made} proposed MADE, a masked feed-forward network that efficiently computes an autoregressive transformation. MADEs are important building blocks in AFs, such as the inverse autoregressive flows (IAFs) introduced by \citet{kingma2016improved}. IAFs are based on trivially invertible affine transformations of the preceding coordinates of the input vector. The parameters of each transformation (a location and a positive scale) are predicted in parallel with a MADE, and therefore IAFs have a lower triangular Jacobian whose determinant is fast to evaluate. IAFs extend the parametric families available for approximate posterior inference in variational inference. Neural autoregressive flow (NAF) by \citet{huang2018neural} extend IAFs by generalizing the bijective transformation to one that can approximate any monotonically increasing function. They have been used both for parametric density estimation and approximate inference. In \S\ref{sec:naf} we explain NAFs in detail and contrast them with our proposed B-NAFs (also, see Figure \ref{fig:architectures}).

\citet{larochelle2011neural} were among the first to employ neural networks for autoregressive density estimation (NADE), in particular, for high-dimensional binary data. Non-linear independent components estimation (NICE) explored the direction of learning a map from high-dimensional data to a latent space with a simpler factorized distribution \citep{dinh2014nice}. \citet{papamakarios2017masked} proposed masked autoregressive flows (MAFs) as a generalization of real non-volume-preserving flows (Real NVP) by \citet{dinh2016density} showing improvements on density estimation.

In this work, we are modelling a discrete normalizing flow since at each transformation a discrete step is made. Continuous normalizing flows (CNF) were proposed by \citet{chen2018neural} and modelled through a network that instead of predicting the output of the transformation predicts its derivative. The resulting transformation is computed using an ODE solver. \citet{grathwohl2018ffjord} further improved such formulation proposing free-form Jacobian of reversible dynamics (FFJORD).

Orthogonal work has been done for constructing powerful invertible function such as invertible $1 \times 1$ convolution (Glow) by \citet{kingma2018glow}, invertible $d \times d$ convolutions \citep{hoogeboom2019emerging}, and invertible residual networks \citep{behrmann2018invertible}. Additionally, \citet{pmlr-v80-oliva18a} investigated different possibilities for the \emph{conditioner} of an autoregressive transformation (e.g., recurrent neural networks).

\begin{figure}[t]
 \centering
 \begin{subfigure}[b]{\fw}
 \centering
 \caption*{Data}
 \vspace{-2mm}
 \includegraphics[width=\linewidth]{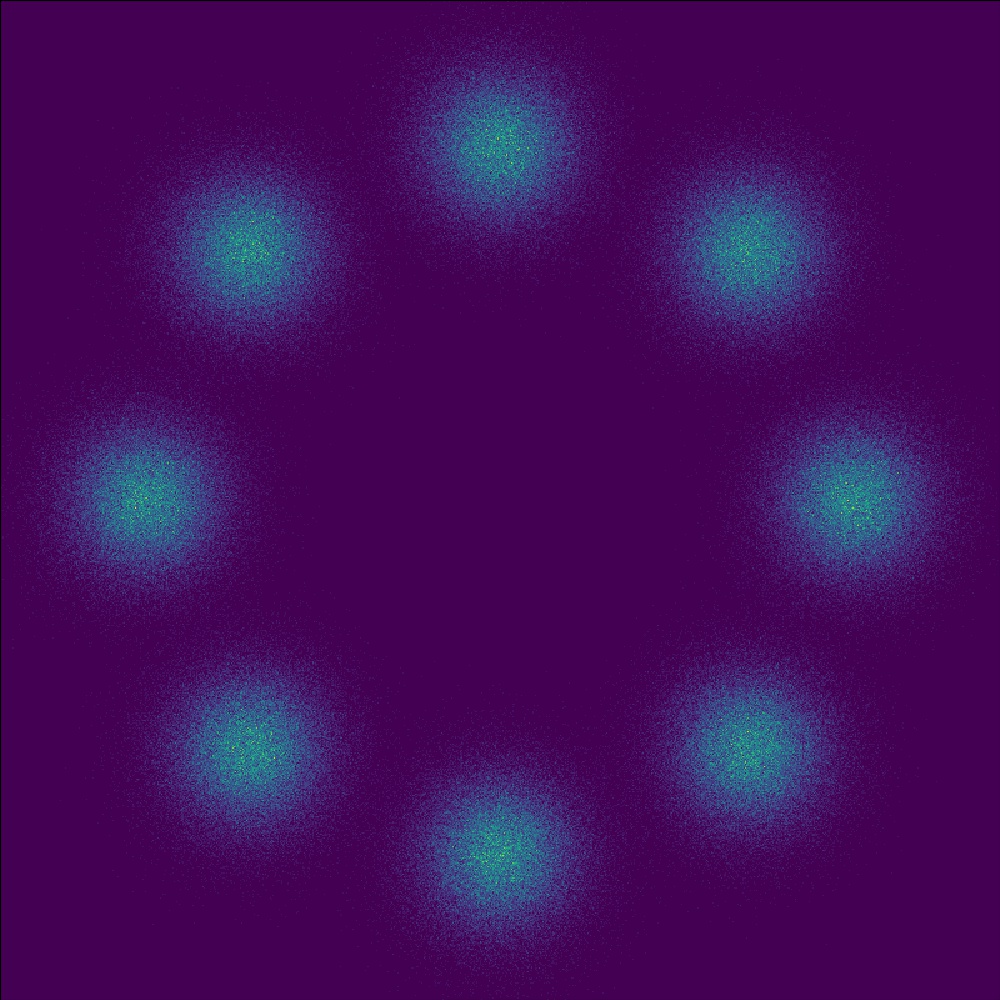}
 \end{subfigure}%
 \hspace{1mm}%
 \begin{subfigure}[b]{\fw}
 \centering
 \caption*{Glow}
 \vspace{-2mm}
 \includegraphics[width=\linewidth]{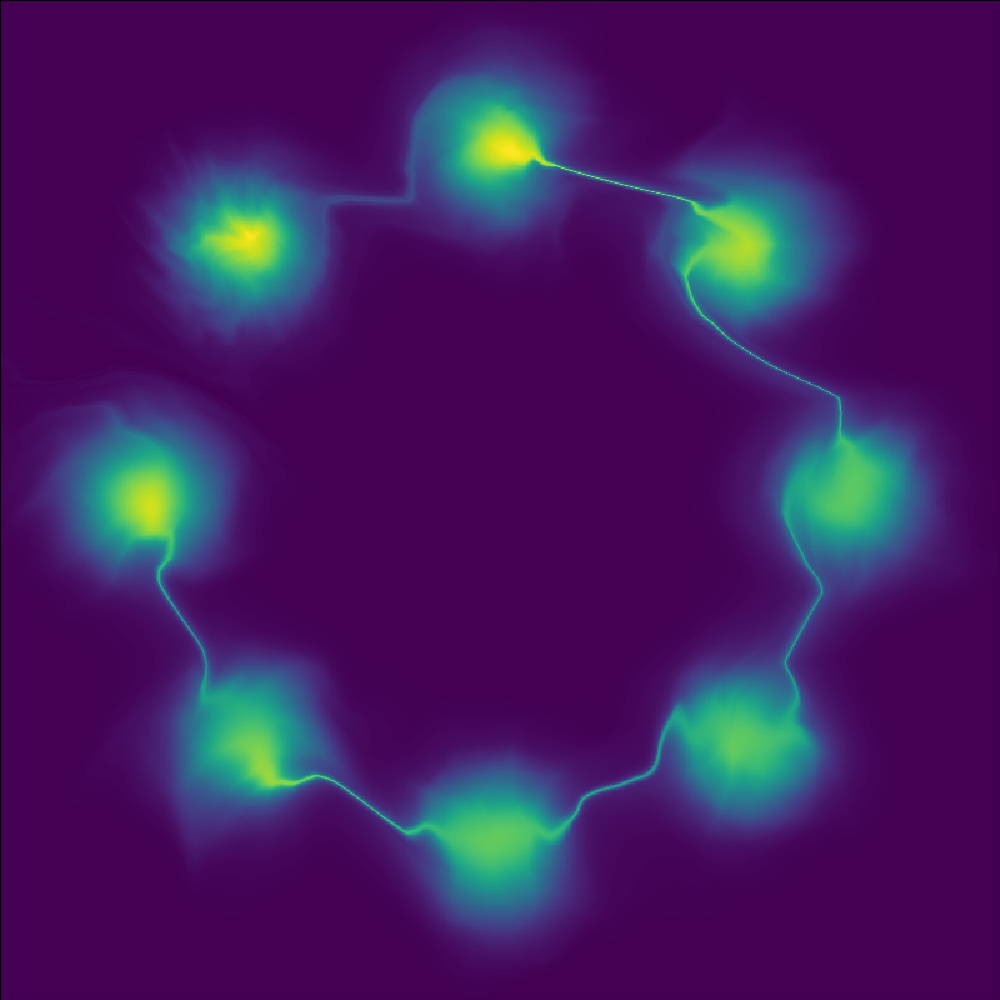}
 \end{subfigure}%
 \hspace{1mm}%
 \begin{subfigure}[b]{\fw}
 \centering
 \caption*{\textbf{Ours}}
 \vspace{-2mm}
 \includegraphics[width=\linewidth]{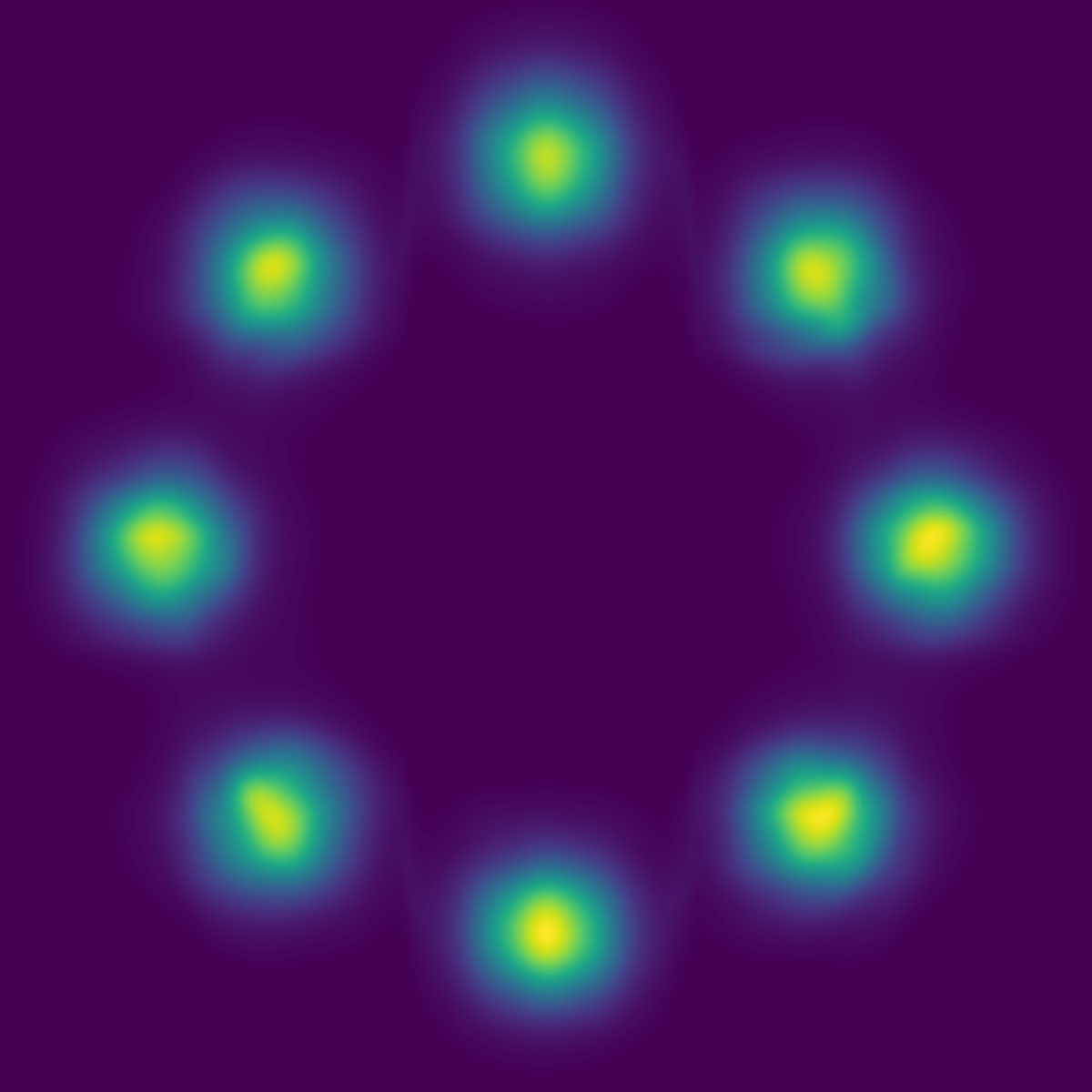}
 \end{subfigure}%
 
 \vspace{1em}
 \begin{subfigure}[b]{\fw}
 \centering
 \vspace{-2mm}
 \includegraphics[width=\linewidth]{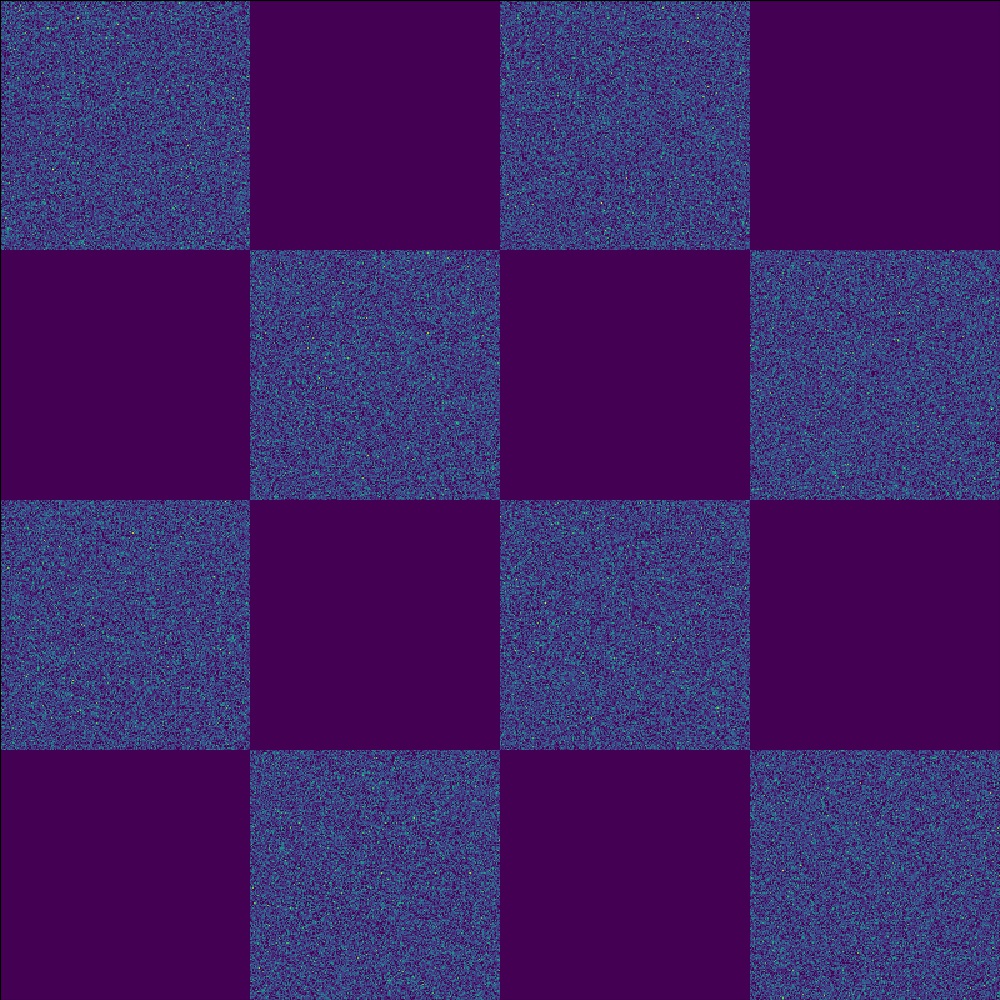}
 \end{subfigure}%
 \hspace{1mm}%
 \begin{subfigure}[b]{\fw}
 \centering
 \vspace{-2mm}
 \includegraphics[width=\linewidth]{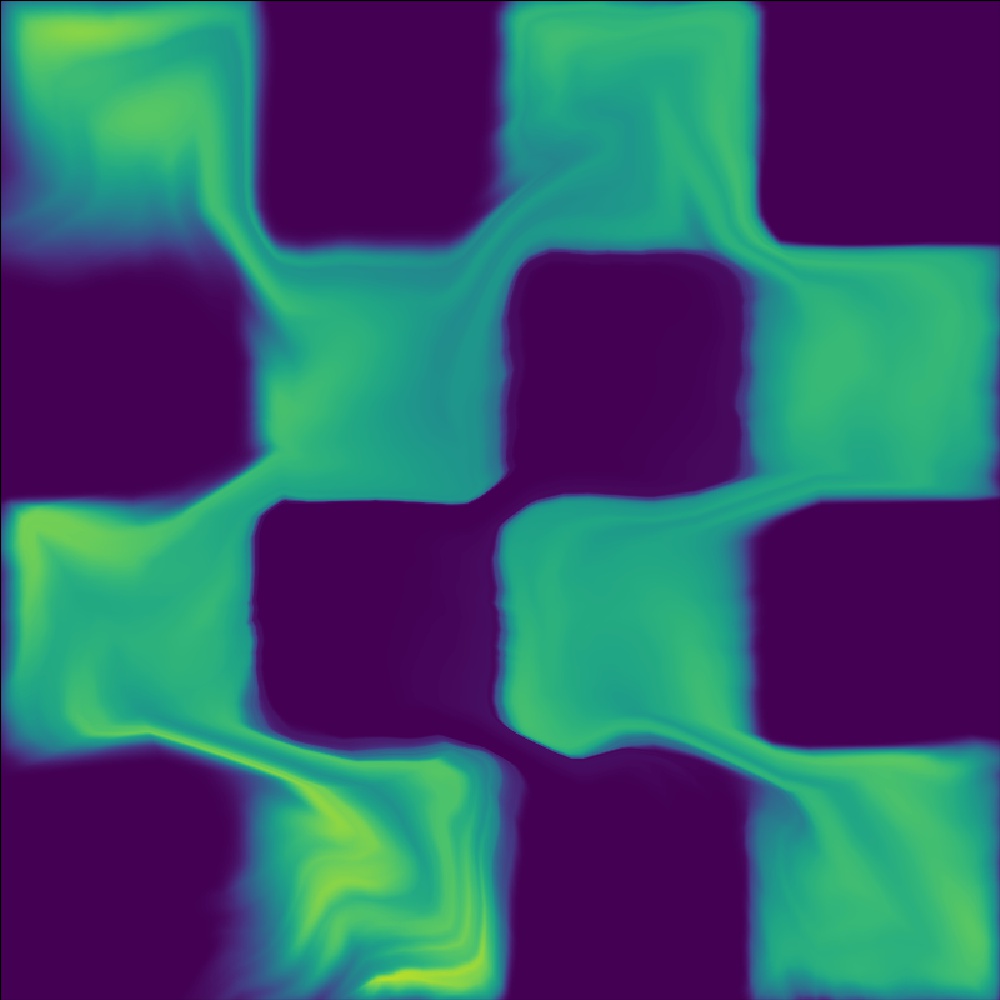}
 \end{subfigure}%
 \hspace{1mm}%
 \begin{subfigure}[b]{\fw}
 \centering
 \vspace{-2mm}
 \includegraphics[width=\linewidth]{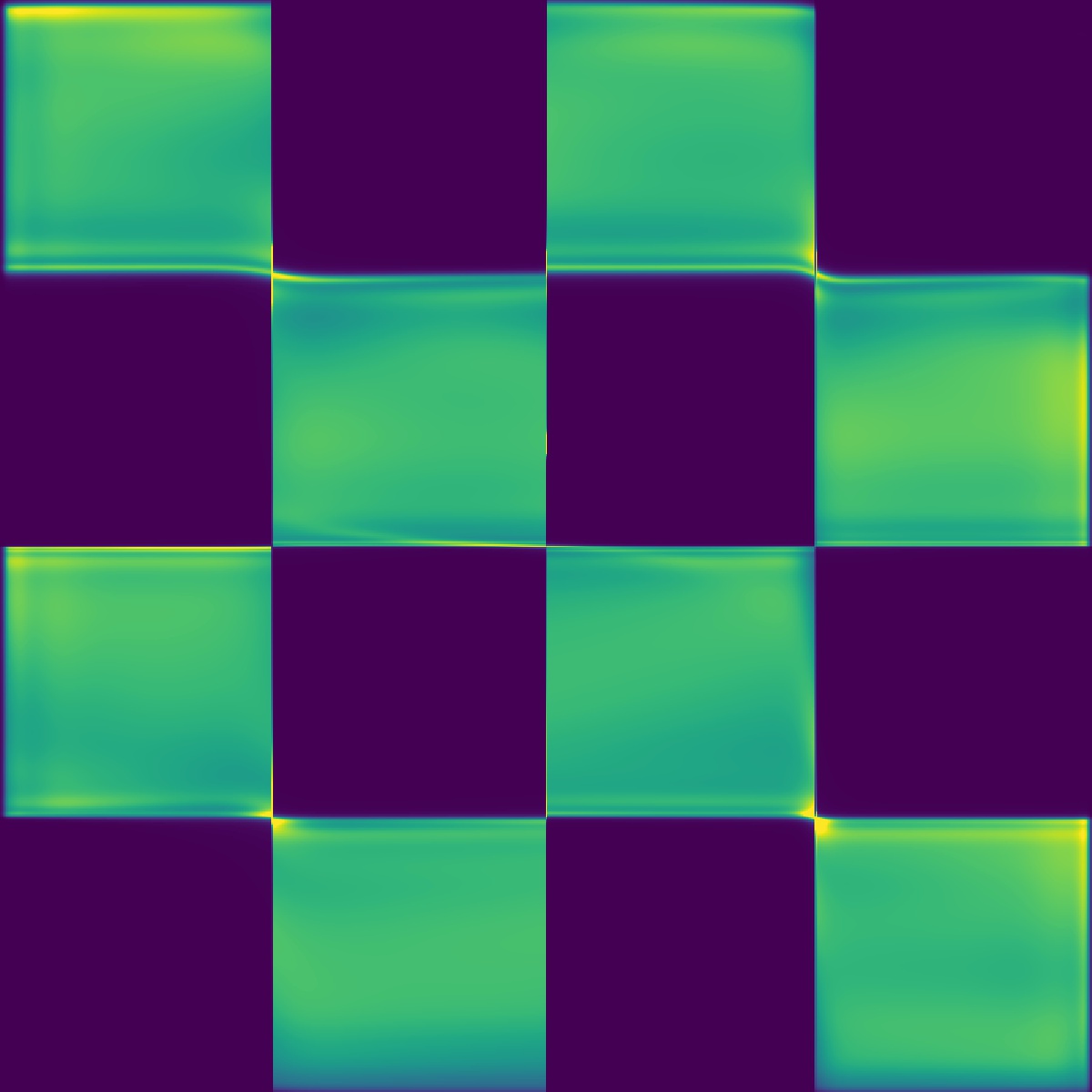}
 \end{subfigure}%
 
 \vspace{1em}
 \begin{subfigure}[b]{\fw}
 \centering
 \vspace{-2mm}
 \includegraphics[width=\linewidth]{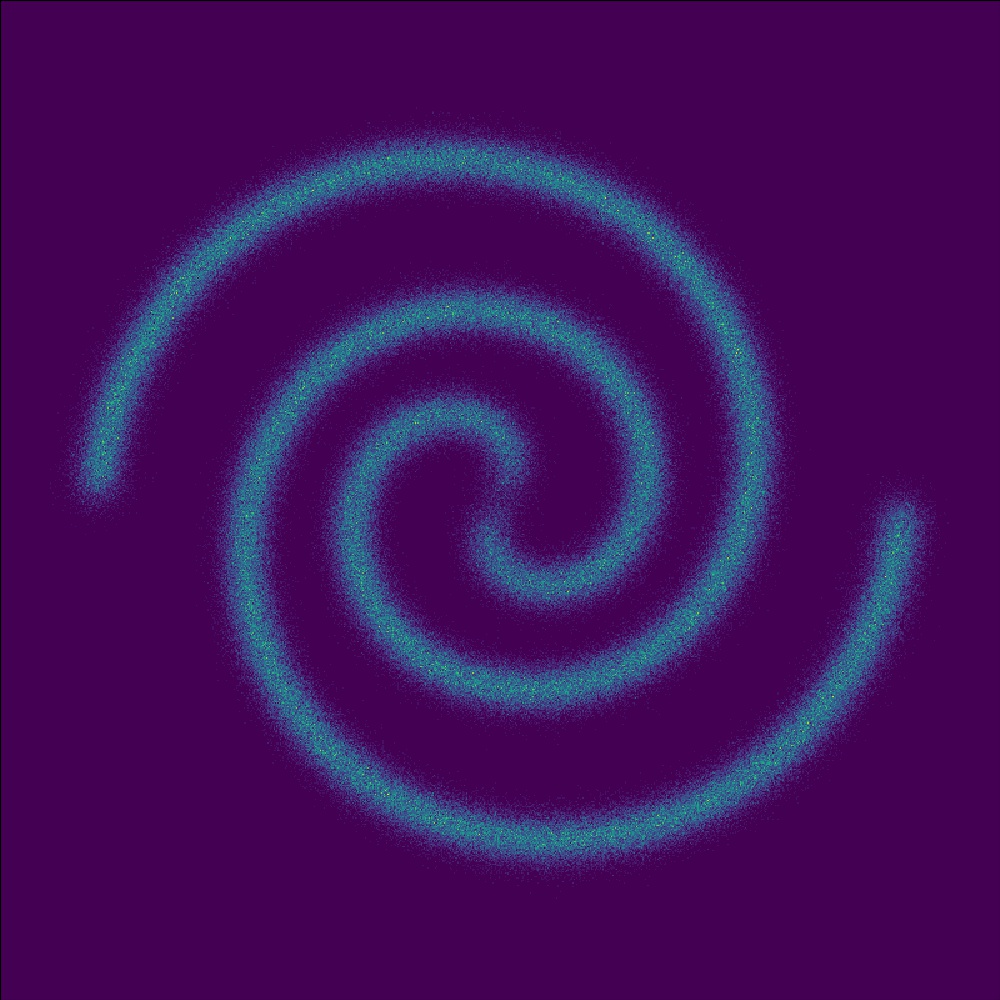}
 \end{subfigure}%
 \hspace{1mm}%
 \begin{subfigure}[b]{\fw}
 \centering
 \vspace{-2mm}
 \includegraphics[width=\linewidth]{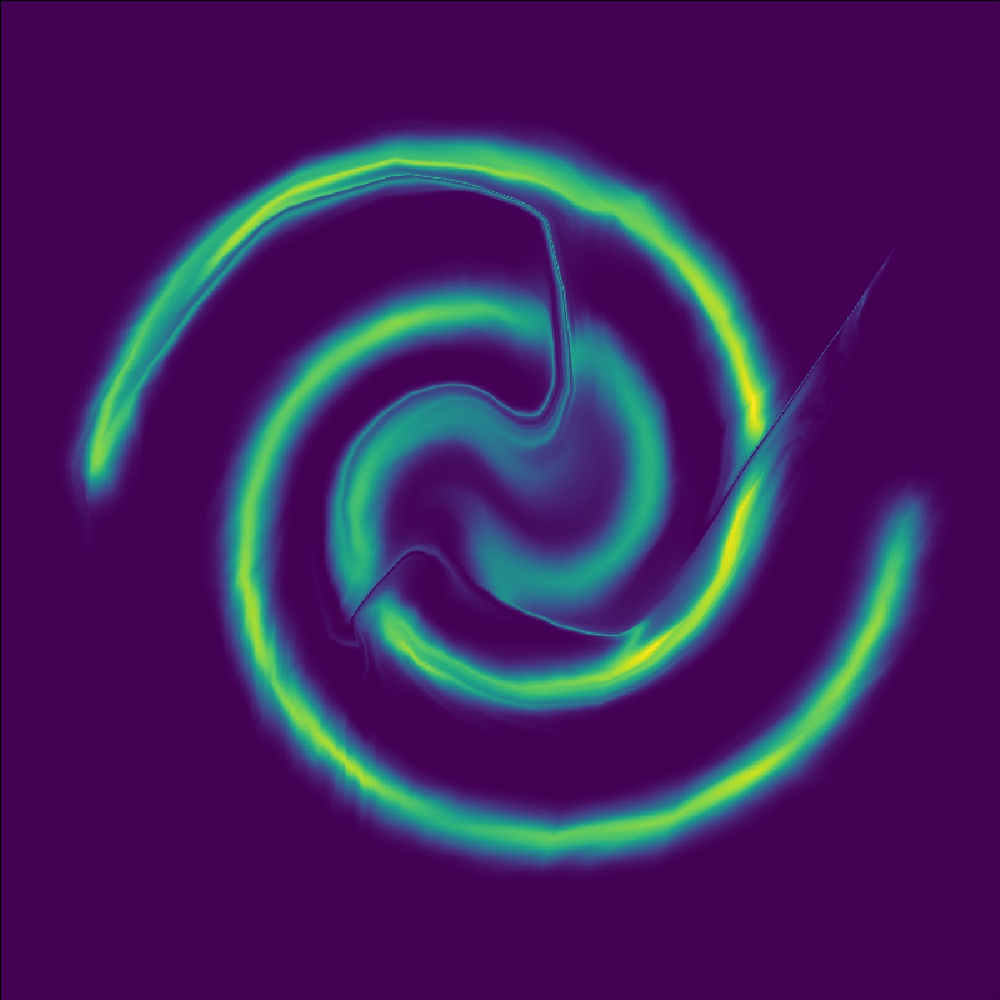}
 \end{subfigure}%
 \hspace{1mm}%
 \begin{subfigure}[b]{\fw}
 \centering
 \vspace{-2mm}
 \includegraphics[width=\linewidth]{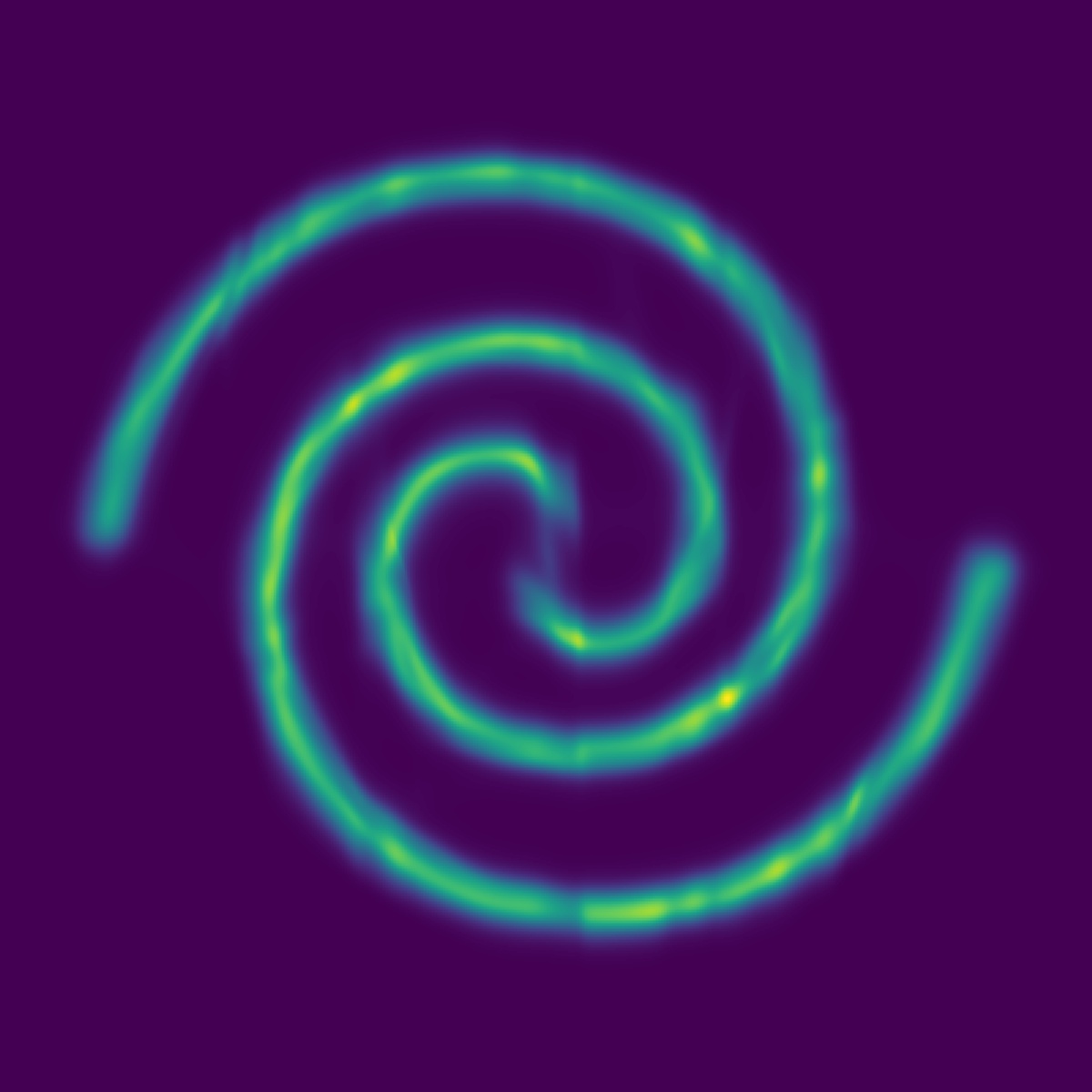}
 \end{subfigure}%
 \caption{Comparison between Glow and B-NAF on density estimation for 2D toy data.}
 \label{fig:toy2d_mle}
\end{figure}

\begin{figure}[t]
 \centering

 \begin{subfigure}[b]{3mm}
 \centering
 1
 \vspace{2em}
 \end{subfigure}%
 \begin{subfigure}[b]{\fw}
 \centering
 \caption*{Target}
 \vspace{-2mm}
 \includegraphics[width=\linewidth]{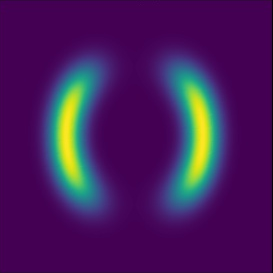}
 \end{subfigure}%
 \hspace{1mm}%
 \begin{subfigure}[b]{\fw}
 \centering
 \caption*{PF (K=32)}
 \vspace{-2mm}
 \includegraphics[width=\linewidth]{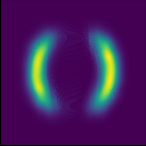}
 \end{subfigure}%
 \hspace{1mm}%
 \begin{subfigure}[b]{\fw}
 \centering
 \caption*{\textbf{Ours}}
 \vspace{-2mm}
 \includegraphics[width=\linewidth]{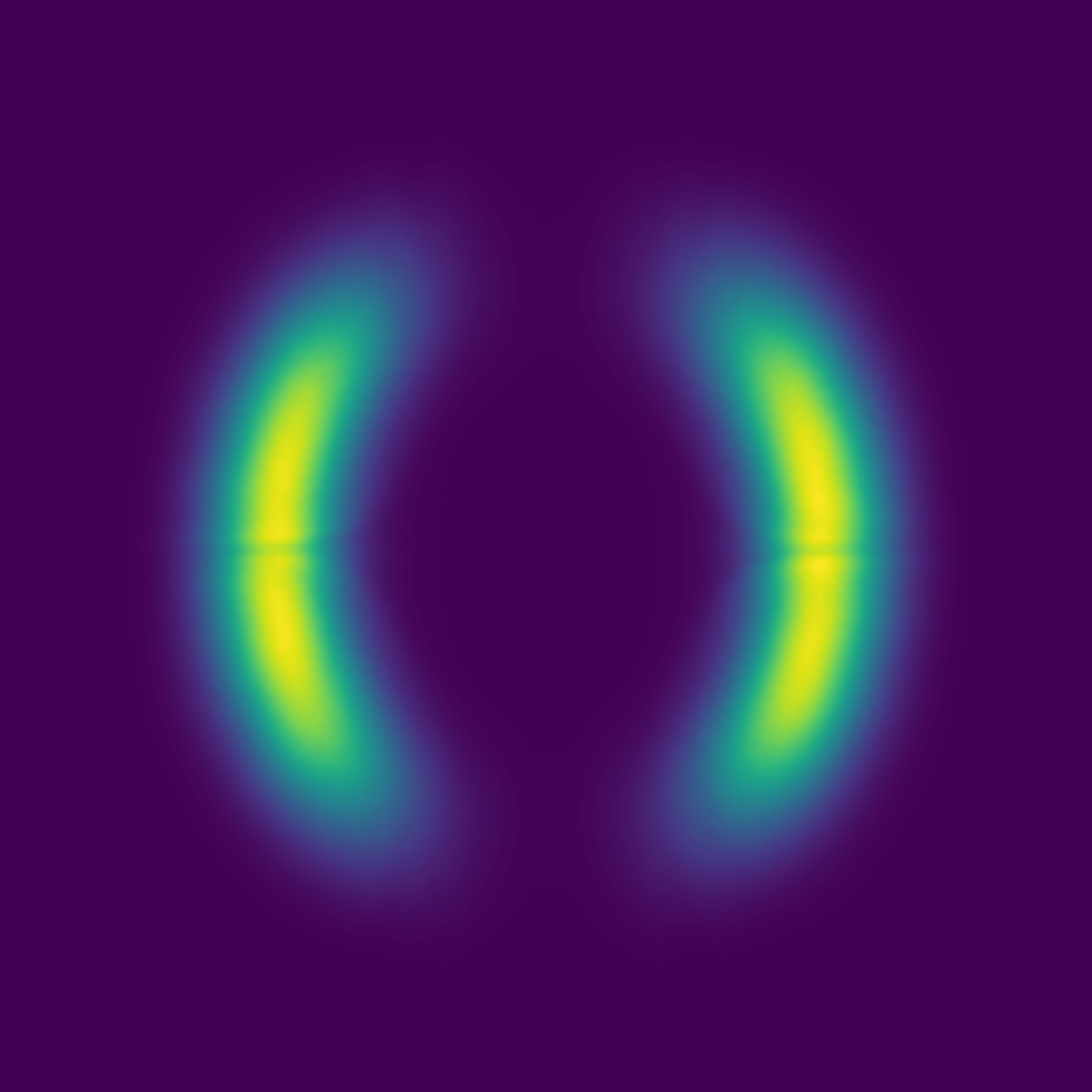}
 \end{subfigure}%
 
 \vspace{1em}
 \begin{subfigure}[b]{3mm}
 \centering
 2
 \vspace{2em}
 \end{subfigure}%
 \begin{subfigure}[b]{\fw}
 \centering
 \vspace{-2mm}
 \includegraphics[width=\linewidth]{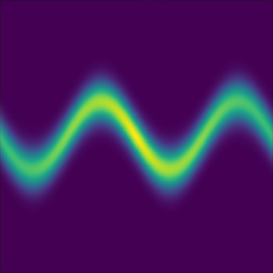}
 \end{subfigure}%
 \hspace{1mm}%
 \begin{subfigure}[b]{\fw}
 \centering
 \vspace{-2mm}
 \includegraphics[width=\linewidth]{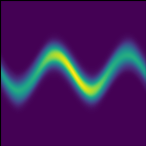}
 \end{subfigure}%
 \hspace{1mm}%
 \begin{subfigure}[b]{\fw}
 \centering
 \vspace{-2mm}
 \includegraphics[width=\linewidth]{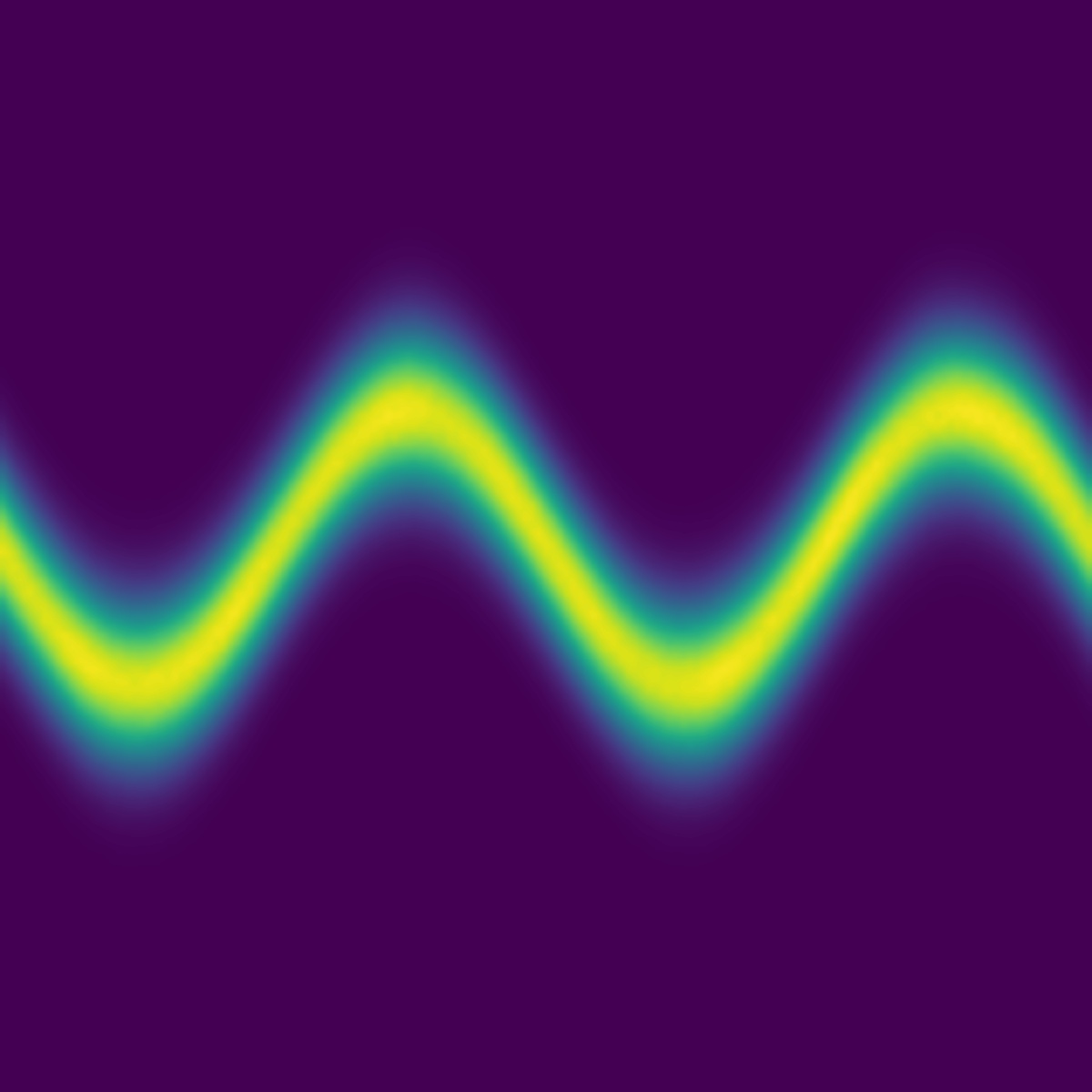}
 \end{subfigure}%
 
 \vspace{1em}
 \begin{subfigure}[b]{3mm}
 \centering
 3
 \vspace{2em}
 \end{subfigure}%
 \begin{subfigure}[b]{\fw}
 \centering
 \vspace{-2mm}
 \includegraphics[width=\linewidth]{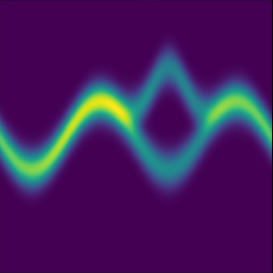}
 \end{subfigure}%
 \hspace{1mm}%
 \begin{subfigure}[b]{\fw}
 \centering
 \vspace{-2mm}
 \includegraphics[width=\linewidth]{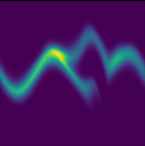}
 \end{subfigure}%
 \hspace{1mm}%
 \begin{subfigure}[b]{\fw}
 \centering
 \vspace{-2mm}
 \includegraphics[width=\linewidth]{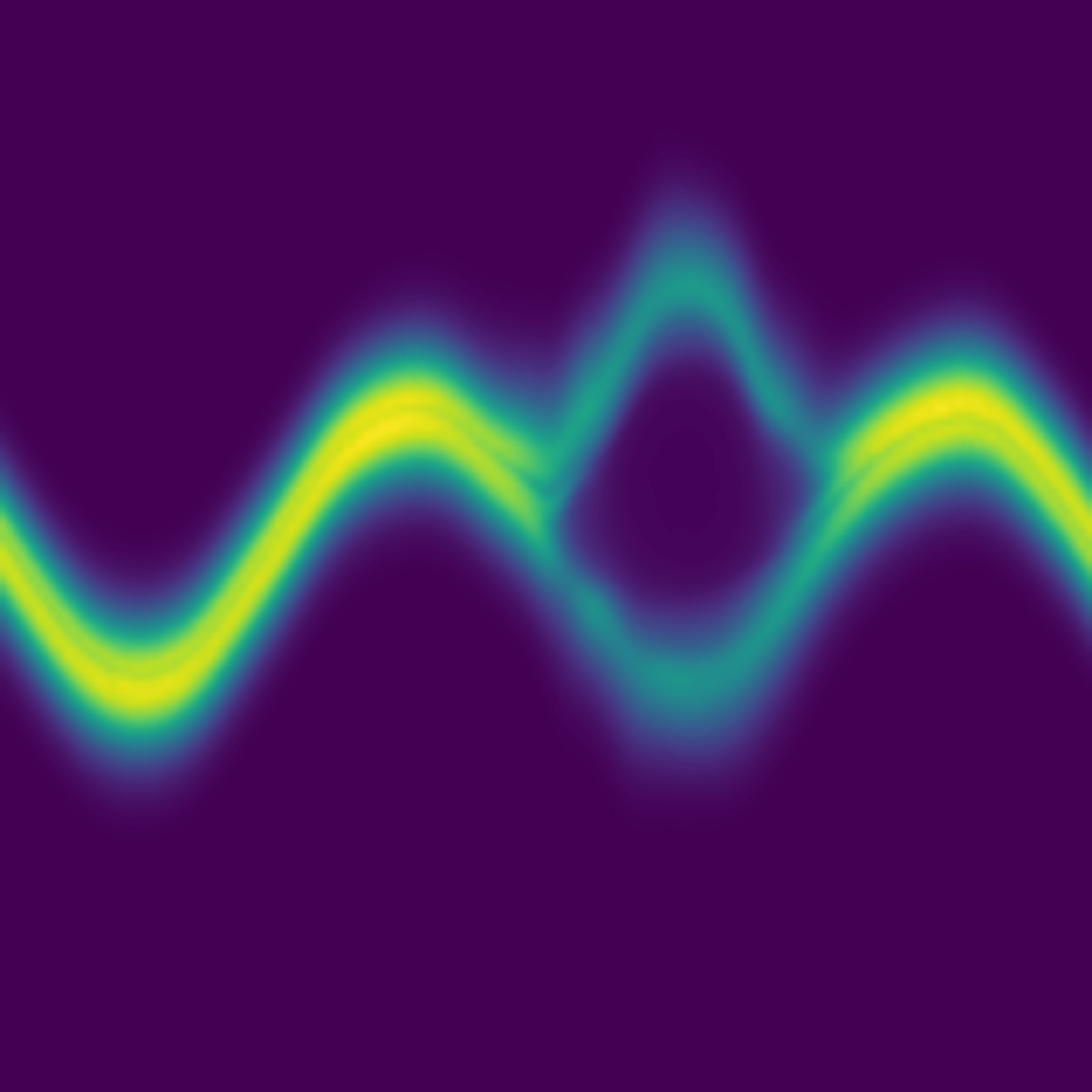}
 \end{subfigure}%
 
 \vspace{1em}
 \begin{subfigure}[b]{3mm}
 \centering
 4
 \vspace{2em}
 \end{subfigure}%
 \begin{subfigure}[b]{\fw}
 \centering
 \vspace{-2mm}
 \includegraphics[width=\linewidth]{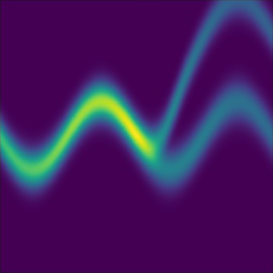}
 \end{subfigure}%
 \hspace{1mm}%
 \begin{subfigure}[b]{\fw}
 \centering
 \vspace{-2mm}
 \includegraphics[width=\linewidth]{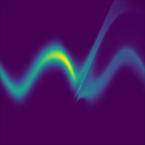}
 \end{subfigure}%
 \hspace{1mm}%
 \begin{subfigure}[b]{\fw}
 \centering
 \vspace{-2mm}
 \includegraphics[width=\linewidth]{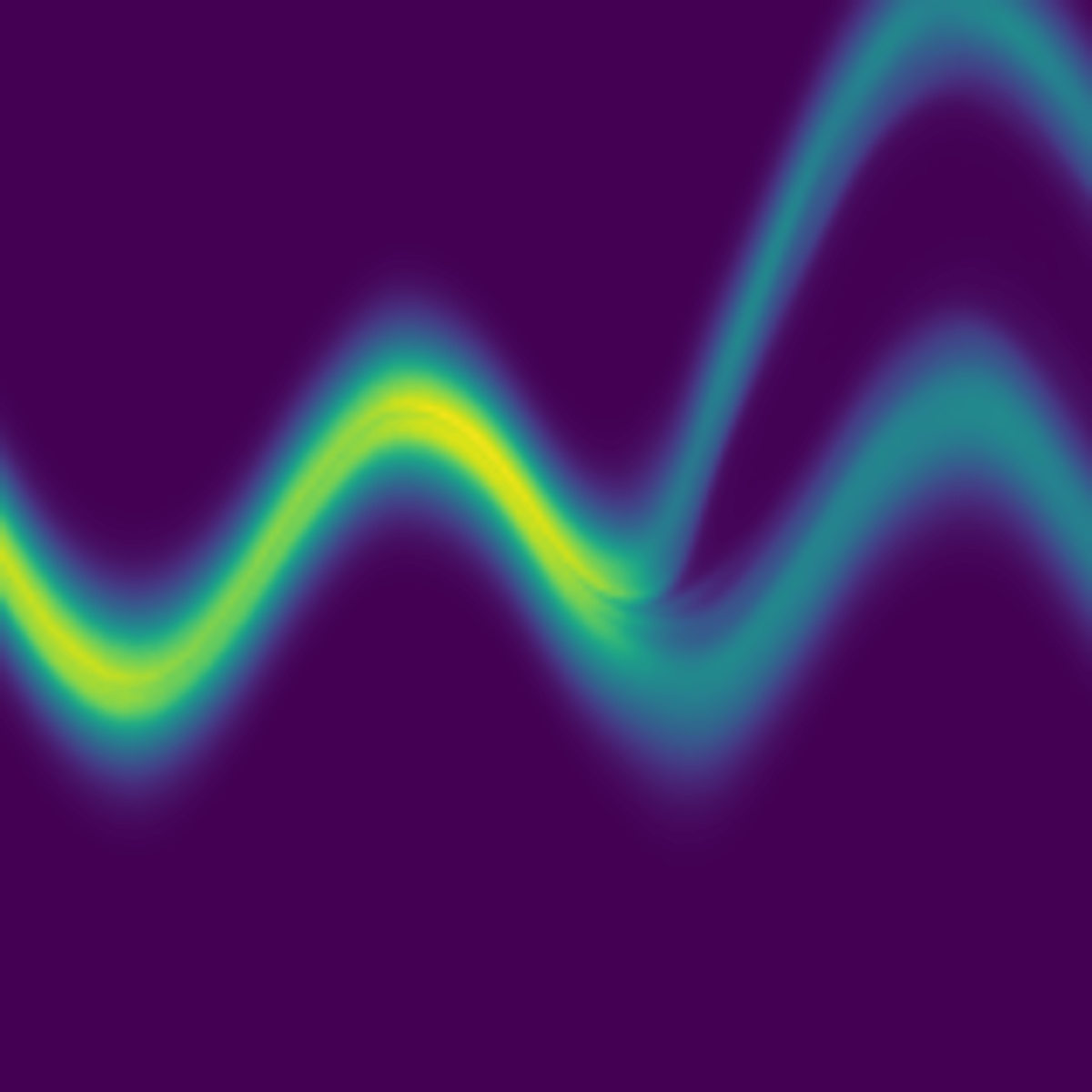}
 \end{subfigure}%
 \caption{Comparison between planar flow (PF) and B-NAF on four 2D energy functions from Table 1 of \citep{rezende2015variational}.}
 \label{fig:toy2d_kl}
\end{figure}

\section{EXPERIMENTS}

\begin{table*}[t]
\centering
\caption{Log-likelihood on the test set (higher is better) for $4$ datasets \citep{dua2017} from UCI machine learning and BSDS300 \citep{martin2001database}. Here $d$ is the dimensionality of datapoints and $N$ the size of the dataset. We report average ($\pm \text{std}$) across $3$ independently trained models. }
\begin{tabular}{lccccc}
\toprule
\multirow{2}{*}{\textbf{Model}} & POWER$\uparrow$ & GAS$\uparrow$ & HEPMASS$\uparrow$ & MINIBOONE$\uparrow$ & BSDS300$\uparrow$ \\
 & {\tiny$d$=6;$N$=2,049,280 } &{\tiny $d$=8;$N$=1,052,065} &{\tiny $d$=21;$N$=525,123} &{\tiny$d$=43;$N$=36,488} & {\tiny$d$=63;$N$=1,300,000} \\
\midrule
Real NVP & $0.17${\tiny$\pm .01$} & $8.33${\tiny$\pm .14$} & $-18.71${\tiny$\pm .02$} & $-13.55${\tiny$\pm .49$} & $153.28${\tiny$\pm 1.78$} \\
Glow & $0.17${\tiny$\pm .01$} & $8.15${\tiny$\pm .40$} & $-18.92${\tiny$\pm .08$} & $-11.35${\tiny$\pm .07$} & $155.07${\tiny$\pm .03$} \\
MADE MoG & $0.40${\tiny$\pm .01$} & $8.47${\tiny$\pm .02$} & $-15.15${\tiny$\pm .02$} & $-12.27${\tiny$\pm .47$} & $153.71${\tiny$\pm .28$} \\
MAF-affine & $0.24${\tiny$\pm .01$} & $10.08${\tiny$\pm .02$} & $-17.73${\tiny$\pm .02$} & $-12.24${\tiny$\pm .45$} & $155.69${\tiny$\pm .28$} \\
MAF-affine MoG & $0.30${\tiny$\pm .01$} & $9.59${\tiny$\pm .02$} & $-17.39${\tiny$\pm .02$} & $-11.68${\tiny$\pm .44$} & $156.36${\tiny$\pm .28$} \\
FFJORD & $0.46${\tiny$\pm .01$} & $8.59${\tiny$\pm .12$} & $-14.92${\tiny$\pm .08$} & $-10.43${\tiny$\pm .04$} & $157.40${\tiny$\pm .19$} \\
NAF-DDSF & $0.62${\tiny$\pm .01$} & $11.96${\tiny$\pm .33$} & $-15.09${\tiny$\pm .40$} & $-8.86${\tiny$\pm .15$} & $157.43${\tiny$\pm .30$} \\
TAN & $0.60${\tiny$\pm .01$} & $12.06${\tiny$\pm .02$} & $-13.78${\tiny$\pm .02$} & $-11.01${\tiny$\pm .48$} & $159.80${\tiny$\pm .07$} \\
\midrule
\textbf{Ours} & $0.61${\tiny$\pm .01$} & $12.06${\tiny$\pm .09$} & $-14.71${\tiny$\pm .38$} & $-8.95${\tiny$\pm .07$} & $157.36${\tiny$\pm .03$} \\
\bottomrule
\end{tabular}
\label{tab:density}
\end{table*}

\subsection{DENSITY ESTIMATION ON TOY 2D DATA} \label{sec:toy2d_mle}
In this experiment, we use our B-NAF to perform density estimation on $2$-dimensional data as this helps us visualizing the model capabilities to learn. We use the same toy data as \citet{grathwohl2018ffjord} comparing the results with Glow \citep{kingma2018glow}, as they do. Given samples from a dataset with empirical distribution $p_{\text{data}}$, we parametrize a density $p_{X|\theta}$ with a normalizing flow $p_{X|\theta}(x) = p_Y(f_\theta(x)) | \det \textbf{J}_{f_\theta(x)} |$ using B-NAF with $p_Y$ a standard Normal distribution. We train for $20$k iterations a single flow of B-NAF with $3$ hidden layers of $100$ units each using maximum likelihood estimation (i.e., maximizing $\mathbb{E}_{\;p_{\text{data}}}[\log p_{X|\theta}(x)]$, see Appendix~\ref{app:kl_desity_estimation} for more details and derivation of the objective). We used Adam optimizer \citep{kingma2014adam} with an initial learning rate of $\alpha=10^{-1}$ (and decay of $0.5$ with patience of $2$k steps), default $\beta_1, \beta_2$, and a batch size of $200$. We took figures of Glow from \citep{grathwohl2018ffjord} who trained such models with $100$ layers.

\paragraph{Results}
The learned distributions of both Glow and our method can be seen in Figure~\ref{fig:toy2d_mle}. Glow is capable of learning a multi-modal distribution, but it has issues assigning the correct density in areas of low probability between disconnected regions. Our model is instead able to perfectly capture both multi-modality and discontinuities.

\subsection{DENSITY MATCHING ON TOY 2D DATA} \label{sec:toy2d_kl}
In this experiment, we use B-NAF to perform density matching on $2$-dimensional target energy functions to visualize the model capabilities of matching them. We use the same energy functions described by \citet{rezende2015variational} comparing the results with them (using planar flows). For this task, we train a parameterized flow minimizing the KL divergence between the learned $q_{Y|\theta}$ and the given target $p_Y$. We used a single flow using a B-NAF with $2$ hidden layers of $100$ units each. We train by minimizing $\KL(q_{Y|\theta} \| p_Y)$ (see Appendix~\ref{app:kl_energy_function} for a detailed derivation) using Monte Carlo sampling. We optimized using Adam for $20$k iterations with an initial learning rate of $\alpha=10^{-2}$ (and decay of $0.5$ with patience of $2$k steps), default $\beta_1, \beta_2$, and a batch size of $200$. Planar flow figures were taken from \citet{chen2018neural}. Note that planar flows were trained for 500k iterations using RMSProp \citep{tieleman2012lecture}.

\paragraph{Results}
Figure~\ref{fig:toy2d_kl} shows that our model perfectly matches all target distributions. Indeed, on functions $3$ and $4$ it looks like B-NAF can better learn the density in certain areas. The model capacity of planar normalizing flows is determined by their depth ($K$) and \citet{rezende2015variational} had to stack $32$ flows to match the energy function reasonably well. Deeper networks are harder to optimize, and our flow matches all the targets using a neural network with only $2$ hidden layers.

\subsection{REAL DATA DENSITY ESTIMATION} \label{sec:density_estimation}
In this experiment, we use a B-NAF to perform density estimation on $5$ real datasets. Similarly to Section~\ref{sec:toy2d_mle}, we train using MLE maximizing $\mathbb{E}_{\;p_{\text{data}}}[\log p_{X|\theta}(x)]$. We compare our results against Real NVP \citep{dinh2016density}, Glow \citep{kingma2018glow}, MADE \citep{germain2015made}, MAF \citep{papamakarios2017masked}, TAN \citep{pmlr-v80-oliva18a}, NAF \citep{huang2018neural}, and FFJORD \citep{grathwohl2018ffjord}. For our B-NAF, we stacked $5$ flows and we employed a small grid search on the number of layers and the size of hidden units per flow ($L \in \{1, 2\}$ and $H \in \{10d, 20d, 40d\}$, respectively, where $d$ is the input size of datapoints which is different for each dataset). When stacking B-NAF flows, the elements of each output vector are permuted so that a different set of elements is considered at each flow. This technique is not novel and it is also used by \citet{dinh2016density, papamakarios2017masked, kingma2016improved}. We trained using Adam with Polyak averaging (with $\phi=0.998$) as in NAF \citep{polyak1992acceleration}. We also applied an exponentially decaying learning rate schedule (from $\alpha=10^{-2}$ with rate $\lambda=0.5$) based on no-improvement with patience of $20$ epochs. We trained until convergence (but maximum $1$k epochs), stopping after $100$ epochs without improvement on validation set.

\paragraph{Datasets} Following \citet{papamakarios2017masked}, we perform unconditional density estimation on four datasets \citep{dua2017} from UCI machine learning repository\footnote{\url{http://archive.ics.uci.edu/ml}} as well as one dataset of patches of images \citep{martin2001database}: POWER containing electric power consumption in a household over a period of $4$ years, GAS containing logs of $8$ chemical sensors exposed to a mixture of gases, HEPMASS, a dataset from a Monte Carlo simulation for high energy physics experiments, MINIBOONE that contains examples of electron neutrino and muon neutrino, and BSDS300 which is obtained by extracting random patches from the homonym datasets of natural images.

\paragraph{Results}
Table~\ref{tab:density} shows the results of this experiment reporting log-likelihood on test set. In all datasets, our B-NAF is better than Real NVP, Glow, MADE, and MAF and it performs comparable or better to NAF. B-NAF also outperforms FFJORD in all dataset except on BSDS300 where there is a marginal difference ($<0.02\%$) between the two methods. On GAS and HEPMASS, B-NAF performs better than most of the other models and even better than NAF. In the other datasets, the gap in performance compared to NAF is marginal. We observed that in most datasets, the best performing model was the largest one in the grid search ($L=2$ and $H=40d$). It is possible that we do a too narrow hyper-parameter search compared to what other methods do. For instance, FFJORD results come from a wider grid search than ours. \citet{grathwohl2018ffjord}, \citet{huang2018neural}, and \citet{pmlr-v80-oliva18a} also varied the number of flows during tuning.

We compare NAF and our B-NAF in terms of the number of parameters employed and report the ratio between the two for each dataset in Table~\ref{tab:params}. For datasets with low-dimensional datapoints (i.e, GAS and POWER) our model uses a comparable number of parameters to NAF. For high-dimensional datapoints the gap between the parameters used by NAF and B-NAF grows, with B-NAF much smaller, as we intended. For instance, on both HEPMASS and MINIBOONE, our models have marginal differences in performance with NAF while having respectively $\sim18\times$ and $\sim40\times$ fewer parameters than NAF. This evidence supports our argument that NAF models are over-parametrized and it is possible to achieve similar performance with an order of magnitude fewer parameters. Besides, when training models on GPUs, being memory efficiency allows to train more models in parallel on the same device. Additionally, in general, a normalizing flow can be a component of a larger architecture that might require more memory than the flow itself (as the models for experiments in the next Section).

\begin{table}[t]
\centering
\caption{Ratios between the number of parameters used by NAF-DDSF (with $5$ or $10$ flows) and our B-NAF on all datasets for density estimation ($d$
is the dimensionality of datapoints). In highly dimensional datasets B-NAF uses orders of magnitude fewer parameters than NAF.}
\begin{tabular}{lrcc}
\toprule
\textbf{Dataset} && NAF (5) & NAF (10) \\
\midrule
POWER & ($d=6$) & $2.29$ & $4.57$ \\
GAS &($d=8$)& $1.30$ & $2.60$ \\
HEPMASS &($d=21$)& $17.94$ & $35.88$ \\
MINIBOONE&($d=43$)& $43.97$ & $87.91$ \\
BSDS300 &($d=63$)& $8.24$ & $16.48$ \\
\bottomrule
\end{tabular}
\label{tab:params}
\end{table}

\begin{table*}[ht]
\centering
\caption{Negative log-likelihood (NLL) and negative evidence lower bound (-ELBO) for static MNIST, Freyfaces, Omniglot and Caltech 101 Silhouettes datasets. For the Freyfaces dataset the results are reported in bits per dim. For the other datasets the results are reported in nats. For all datasets we report the mean and the standard deviations over $3$ runs with different random initializations.}

\begin{tabular}{lcccccccc}
\toprule
\multirow{2}{*}{\bfseries{Model}} & \multicolumn{2}{c}{\bfseries MNIST} & \multicolumn{2}{c}{\bfseries Freyfaces} & \multicolumn{2}{c}{\bfseries Omniglot} & \multicolumn{2}{c}{\bfseries Caltech 101} \\
& \multicolumn{1}{c}{-ELBO$\downarrow$} & \multicolumn{1}{c}{NLL$\downarrow$} & \multicolumn{1}{c}{-ELBO$\downarrow$} & \multicolumn{1}{c}{NLL$\downarrow$} & \multicolumn{1}{c}{-ELBO$\downarrow$} &\multicolumn{1}{c}{NLL$\downarrow$} & \multicolumn{1}{c}{-ELBO$\downarrow$} &\multicolumn{1}{c}{NLL$\downarrow$} \\
\midrule
VAE & $86.55${\tiny$\pm .06$} & $82.14${\tiny$\pm .07$} & $4.53${\tiny$\pm.02$} & $4.40${\tiny$\pm .03$} & $104.28${\tiny$\pm .39$} & $97.25${\tiny$\pm .23$} & $110.80${\tiny$\pm .46$} & $99.62${\tiny$\pm .74$} \\
Planar & $86.06${\tiny$\pm .31$} & $81.91${\tiny$\pm .22$} & $4.40${\tiny$\pm .06$} & $4.31${\tiny$\pm .06$} & $102.65${\tiny$\pm .42$} & $96.04${\tiny$\pm .28$} & $109.66${\tiny$\pm .42$} & $98.53${\tiny$\pm .68$}\\
IAF & $84.20${\tiny$\pm .17$} & $80.79${\tiny$\pm .12$} & $4.47${\tiny$\pm .05$} & $4.38${\tiny$\pm .04$} & $102.41${\tiny$\pm .04$} & $ 96.08${\tiny$\pm .16$} & $111.58${\tiny$\pm .38$} & $ 99.92${\tiny$\pm .30$}\\
Sylvester & $83.32${\tiny$\pm .06$} & $80.22${\tiny$\pm .03$} & $4.45${\tiny$\pm .04$} & $4.35${\tiny$\pm .04$}& $99.00${\tiny$\pm .04$} & $93.77${\tiny$\pm .03$} & $104.62${\tiny$\pm .29$} & $93.82${\tiny$\pm .62$}\\
\midrule
\textbf{Ours} & $83.59${\tiny$\pm .15$} & $80.71${\tiny$\pm .09$} & $4.42${\tiny$\pm .05$} & $4.33${\tiny$\pm .04$} & $100.08${\tiny$\pm .07$} & $94.83${\tiny$\pm .10$} & $105.42${\tiny$\pm .49$} & $94.91${\tiny$\pm .51$} \\
\bottomrule
\end{tabular}
\label{tab:vae}
\end{table*}

\subsection{VARIATIONAL AUTO-ENCODERS} \label{sec:vae}
An interesting application of our framework is modelling more flexible posterior distributions in a variational auto-encoder (VAE) setting \citep{kingma2013auto}. In this setting, we assume that an observation $x$ (i.e., the data) is drawn from the marginal of a deep latent model, i.e. $X \sim p_{X|\theta}$, where $p_{X|\theta}(x) = \int p_Z(z) p_{X|Z,\theta}(x|z) \dd z$ where $Z \sim \mathcal N(0, I)$ is unobserved. The goal is performing maximum likelihood estimation of the marginal. Since $Z$ is not observed, maximizing the objective would require marginalization over the latent variables, which is generally intractable. Using variational inference \citep{jordan1999introduction}, we can maximize a lower bound on log-likelihood:
\begin{equation} \label{eq:elbo}
 \log p_{X|\theta}(x) \geq \mathbb E_{\;q_{Z|X,\phi}(z)} \left[ \log \frac{p_{XZ|\theta}(x, z)}{q_{Z|X,\phi}(z|x)} \right] \;,
\end{equation}
where $p_{X|Z,\theta}$ and $q_{Z|X,\phi}$ are parametrized via neural networks with learnable parameters $\theta$ and $\phi$ \citep{kingma2013auto}, in particular, $q_{Z|X, \phi}$ is an approximation to the intractable posterior $p_{Z|X,\theta}$. 
This bound is called the evidence lower bound (ELBO), and maximizing the ELBO is equivalent to minimizing $\KL(q_{Z|X, \phi} \| p_{Z|X,\theta})$. The more expressive the approximating family is, the more likely we are to obtain a tight bound. Recent literature approaches tighter bounds by approximating the posterior with normalizing flows. Also note that NFs reparametrize $q_{Z|X,\phi}(z|x) = q_{Y}(f_\phi(z; x)) \abs{\det \mathbf J_{f_\phi(z; x)}}$ via a simpler fixed base distribution, e.g. a standard Gaussian, and therefore we can follow stochastic gradient estimates of the ELBO with respect to both sets of parameters. In this experiment, we use our flow for posterior approximation showing that B-NAF compares with recently proposed NFs for variational inference. We reproduce experiments by \citet{berg2018sylvester} (Sylvester flows or SNF) while replacing their flow with ours. We keep the encoder and decoder networks exactly the same to fairly compare with all models trained with such procedure. We compare our B-NAF to their flows on the same $4$ datasets as well as to a normal VAE \citep{kingma2013auto}, planar flows \citep{rezende2015variational}, and IAFs \citep{kingma2016improved}.\footnote{
Although also \citet{huang2018neural} proposed an experiment with VAEs for NAF, they used only one dataset (MNIST) employed a different encoder/decoder architecture than \citet{berg2018sylvester}. Therefore, results are not comparable.}

In this experiment, the input dimensionality of the flow is fixed to $d=64$. We employed a small grid search on the MNIST dataset on the number of flows $K \in \{4, 8\}$, and on thee size of hidden units per flow $H \in \{2d, 4d, 8d\}$ while keeping the number of layers fixed at $L=1$. The elements of each output vector are permuted after each B-NAF flow (as we do in \S~\ref{sec:density_estimation}). We keep the best hyper-parameters of this search for the other datasets. We train using Adamax with $\alpha=5 \cdot 10^{-4}$. We point to Appendix A of \citet{berg2018sylvester} for details on the network architectures for the encoder and decoder.

\paragraph{Datasets} 
Following \citet{berg2018sylvester} we carried our experiments on $4$ datasets: statically binarized MNIST \citep{larochelle2011neural}, Freyfaces\footnote{\url{http://www.cs.nyu.edu/~roweis/data/frey_rawface.mat}}, Omniglot \citep{lake2015human} and Caltech 101 Silhouettes \citep{marlin2010inductive}. All those datasets consist of black and white images of different sizes.

\paragraph{Amortizing flow parameters}
When using NFs in an amortized inference setting, the parameters of each flow are not learned directly but predicted with another function from each datapoint \citep{rezende2015variational}. In our case, we do not amortize all parameters of B-NAF since that would require very large predictors and we want to keep our flow memory efficient. Alternatively, every affine matrix $W \in \mathbb{R}^{n\times m}$ is shared among all datapoints. Then, for each affine transformation, we achieve a degree of amortization by predicting $3$ vectors, the bias $b \in \mathbb{R}^n$ and $2$ vectors $v_1 \in \mathbb{R}^n$ and $v_2 \in \mathbb{R}^m$ that we multiply row- and column-wise respectively to $W$.

\paragraph{Results}
Table~\ref{tab:vae} shows the results of these experiments. From the grid search, it turned out that the best B-NAF model has $K=8$ (flows) and $H=4d$ (hidden units). Note that the best models reported by \citet{berg2018sylvester} used $16$ flows. Our model is quite flexible without being as deep as other models. Results show that B-NAF is better than normal VAE, planar flows, and IAFs in all four datasets. Although B-NAF performs slightly worse than Sylvester flows, \citet{berg2018sylvester} applied a full amortization for the parameters of the flow, while we do not. They proposed two alternative parametrizations to construct Sylvester flows: orthogonal SNF and Houseolder SNF. For each datapoint, SNF has to predict from $50.7$k to $76.8$k values (depending on the parametrization) to fully amortize parameters of the flow, while we use only $7.7$k (i.e., $6.64\times$ to $10.0\times$ fewer). Notably, recalling that these are not trainable parameters, we use $6.16 \times$ (orthogonal SNF) and $9.35\times$ (Householder SNF) fewer trainable parameters as well. Besides, we also use $14.45\times$ fewer parameters than IAF. This shows that IAF and SNF are over-parametrized too, and it is possible to achieve similar performance in the context of variational inference with an order of magnitude fewer parameters.

\section{CONCLUSIONS}
We present a new family of flexible normalizing flows, \emph{block neural autoregressive flows}. B-NAFs are universal approximators of density functions and maintain an efficient parametrization. Our flow is based on directly parametrizing a transformation that guarantees autoregressiveness and monotonicity without the need for large conditioner networks and without compromising parallelism. Compared to the only other flow (to the best of our knowledge) which is also a universal approximator, our B-NAFs require orders of magnitudes fewer parameters. We validate B-NAFs on parametric density estimation on toy and real datasets, as well as, on approximate posterior inference for deep latent variable models, showing favorable performance across datasets and against various established flows. For future work, we are interested in at least two directions. One concerns gaining access to the inverse of the flow---note that, while B-NAFs and NAFs are invertible in principle, their inverses are not available in closed form. Another concerns deep generative models with large decoders (e.g. in natural language processing applications): since we achieve high flexibility at a low memory footprint our flows seem to be a good fit.

\subsection*{Acknowledgements}
We would like to thank George Papamakarios and Luca Falorsi for insightful discussions. This project is supported by SAP Innovation Center Network, ERC Starting Grant BroadSem (678254) and the Dutch Organization for Scientific Research (NWO) VIDI 639.022.518. Wilker Aziz is supported by the European Union's Horizon 2020 research and innovation programme under grant agreement No 825299 (Gourmet).

\bibliography{main}
\bibliographystyle{apalike}
\clearpage

\onecolumn
\appendix

\section{OBJECTIVE FOR DENSITY ESTIMATION} \label{app:kl_desity_estimation}
When performing density estimation for a random variable $X$, we only have access to samples from the unknown target distribution $X \sim p_{\star}$ (i.e., the unknown data distribution) but we do not have access to $p_\star$ directly \citep{papamakarios2017masked}. Using Equation~\ref{eq:flow}, we can use a normalizing flow to transform a complex parametric model $p_{X|\theta}$ of the target distribution into a simpler distribution $p_Y$ (i.e., a uniform or a Normal distribution), which can be easily evaluated. In this case, we will learn the parameters $\theta$ of the model by minimizing $\KL(p_\star \| p_{X|\theta})$:
\begin{align}
 \theta^* &= \min_{\theta} ~ \KL(p_\star \| p_{X|\theta}) \\
 &= \min_{\theta} ~ \mathbb{E}_{p_\star(x)} \left[\log \frac{p_\star(x)}{p_{X|\theta}(x)} \right] \\
 &= \min_{\theta} ~ \underbrace{\mathbb{E}_{p_\star(x)} [\log p_\star(x)]}_{\text{=constant}} - \mathbb{E}_{p_\star(x)}[\log p_{X|\theta}(x)] \\
 &= \max_\theta ~ \mathbb{E}_{p_\star(x)}\left[ \log p_Y(f_\theta(x)) + \log \abs{ \det \mathbf{J}_{f_\theta(x)} } \right] \;.
\end{align}
where $p_{X|\theta}(x) = p_Y(y) \abs{ \det \mathbf{J}_{f_\theta(x)} }$ and $y = f_\theta(t)$. Notice that minimizing the KL is equivalent of doing maximum likelihood estimation (MLE).

\section{OBJECTIVE FOR DENSITY MATCHING} \label{app:kl_energy_function}
We can learn how to sample from a complex target distribution $p_\star$ (or, more generally, an energy function) for which we have access to its analytical form but we do not have an available sampling procedure. Using Equation~\ref{eq:flow}, we can use a normalizing flow to transform samples from a simple distribution $p_X$, which we can easily evaluate and sample from, to a complex one (the target). In this case, we estimate $\theta$ by minimizing $\KL(p_{Y|\theta}\|p_\star)$:
\begin{align}
 \theta^* &= \min_{\theta} ~ \KL(p_{Y|\theta}\|p_\star) \\
 &= \min_{\theta} ~ \mathbb{E}_{p_{Y|\theta}(y)} \left[\log \frac{p_{Y|\theta}(y)}{p_\star(y)} \right] \\
 &= \min_{\theta} ~ \mathbb{E}_{p_{Y|\theta}(y)} [\log p_{Y|\theta}(y) - \log p_\star(y)] \\
 &= \min_\theta ~ \mathbb{E}_{p_X(x)}[\log p_X(x) - \log \abs{ \det \mathbf{J}_{f_\theta(x)} } - \log p_\star(f_\theta(x)) ] \;.
\end{align}
where $p_{Y|\theta}(y) = p_X(x) \abs{ \det \mathbf{J}_{f_\theta(x)} }^{-1}$ and $y = f_\theta(x)$.
Notice that in general, with normalizing flows, it is possible to learn a flexible distribution from which we can sample and evaluate the density of its samples. These two proprieties are particularly useful in the context of variational inference \citep{rezende2015variational}. 

\section{WEIGHT INITIALIZATION AND NORMALIZATION} \label{app:weight}
Since the weight matrix $W$ has some strictly positive and some zero entries, we need to take care of a proper initialization. Indeed, it is well known that principled parameters initialization benefits not only training but also the generalization of neural networks \citep{glorot2010understanding}. For instance, Xavier initialization is commonly used and it takes into account the size of the input and output spaces in the affine transformations. However, since we have some zero entries, we cannot benefit from it. We choose instead to initialize all blocks with a simple distribution and to apply weight normalization \citep{salimans2016weight} to better regulate the effect of such initialization. Weight normalization decomposes each row $w \in \mathbb{R}^{b \cdot d}$ of $W$ in terms of the new parameters using $w = \exp(s) \cdot v/\|v\|$ where $v$ has the same dimensionality of $w$ and $s$ is a scalar. We initialize $v$ with a simple Normal distribution of zero mean and unit variance and $s=\log(u)$ with $u\sim \mathcal{U}(0, 1)$. Such reparametrization disentangles the direction and magnitude of $w$ and it is known to improve and speed up optimization.

\hfill
\clearpage

\end{document}